\documentclass{article}
\usepackage[margin=1.3in]{geometry}
\usepackage{amsmath,amssymb,amsthm}
\usepackage{caption}
\usepackage{booktabs}
\usepackage{graphicx}
\usepackage{hyperref}
\usepackage{url}
\usepackage{microtype}
\usepackage{enumitem}
\usepackage{longtable}
\usepackage{algorithm}
\usepackage{algpseudocode}
\usepackage{booktabs}
\usepackage{multirow}
\usepackage{graphicx}
\usepackage{makecell}
\usepackage{tikz}
\usepackage{xcolor}
\colorlet{blue}{black}
\usepackage{placeins} 
\usepackage{etoolbox} 

\usepackage[numbers]{natbib}

\theoremstyle{remark}

\newcommand{\rt}[1]{#1}
\graphicspath{{figs/}}

\title{DEFTS: Data-Efficient, Training-Free Few-Shot Learning for CLIP in the Text Subspace}
\author{Yijin Zeng\\Independent researcher\\London, United Kingdom\\\texttt{yijinzeng98@gmail.com}}
\date{November 15, 2026}

\begin{document}
\maketitle

\begin{abstract}
Contrastive Language-Image Pre-training (CLIP) achieves impressive zero-shot image 
classification performance by comparing image embeddings against class-specific 
text embeddings. 
While recent few-shot methods leverage labeled training samples to 
improve upon this zero-shot baseline, they often predominantly rely on image embeddings 
and underutilize the semantic structure of the text embeddings. 
In this paper, we introduce DEFTS, a data-efficient, training-free few-shot classification method
strictly in the text subspace.
We first demonstrate empirically that while raw text embeddings can be suboptimal as direct classifiers, the linear subspace they span remains highly discriminative. DEFTS therefore proposes to reduce the image-text modality gap by projecting estimated image prototypes directly into this subspace. Importantly, by expressing these prototypes as linear combinations of text embeddings, DEFTS reduces a high-dimensional estimation problem to the derivation of just a few coefficients. These coefficients are then computed via a Bayesian posterior mean, using a prior naturally centered on the original text embedding coordinates.
Extensive experiments across 12 image classification benchmarks demonstrate that DEFTS significantly improves upon existing training-free methods.
\end{abstract}

\section{Introduction}
\label{sec:intro}

Contrastive Language-Image Pre-training (CLIP) \citep{radford2021learning} 
has gained considerable attention due to its impressive zero-shot capabilities. 
The model is trained to maximize the similarity between matched image-text pairs while
minimizing the similarity across all incorrect pairings. By leveraging over 400 million 
image-text pairs, CLIP shows remarkable zero-shot image classification performance 
without requiring any dataset-specific fine-tuning. This is achieved by passing an image 
through a image encoder and class-specific text prompts (e.g., ``a photo of a \{class\}'') 
through a text encoder, projecting both into a shared semantic space. The image is then 
assigned to the class whose text embedding yields the highest cosine similarity.

Despite its powerful zero-shot performance, recent work 
\citep{wang2024hardtobeat, zhang2022tipadapter, guo2023calip, wang2019simpleshot, udandarao2023susx, zanella2024low, goswami2026cross, zhou2022learning, zhao2024test} has demonstrated 
that the performances of CLIP could be improved significantly by utilizing only a few
labeled training samples, known as few-shot learning.
In particular, training-free few-shots methods \citep{wang2024hardtobeat, zhang2022tipadapter, udandarao2023susx, goswami2026cross} are becoming increasingly more popular because of their computational efficiency and comparable power compared with training-required methods \citep{wang2024hardtobeat}.

While significantly improving upon zero-shot baselines, most training-free methods rely heavily on the image embeddings and underutilize the semantic structure of the text embeddings. For example, \citet{wang2024hardtobeat} fit a Linear Discriminant Analysis (LDA) model directly on the visual features to classify images during inference. Similarly, \citet{zhang2022tipadapter} cache the $k$-shot image embeddings as keys and their one-hot labels as values to compute attention weights for test images. Both approaches only utilize the text embeddings by ensembling with the original zero-shot CLIP predictions.

Therefore, in this paper, we aim to maximize the utility of the text embeddings by constraining the alignment of image and text representations strictly within the text-spanned subspace. Because our approach operates entirely within the the text space, it naturally complements existing image-centric adaptation methods.

\paragraph{Related Work}
Adapting foundational vision-language models to downstream tasks without gradient-based optimization has become highly desirable due to its computational efficiency. Early approaches focused on retrieval and attention mechanisms: \citet{zhang2022tipadapter} build a key-value cache directly from few-shot support features, \citet{udandarao2023susx} impvoes CLIP performances by creating a curated support set, \citet{guo2023calip} apply a parameter-free cross-modal attention module to sharpen zero-shot logits, and \citet{zhu2024selective} align the extracted image and text features through two subspace projections decided by image-image and text-image attention weights.

Statistical classifiers operating directly on the extracted embeddings have also proven remarkably effective. \citet{wang2019simpleshot} demonstrate that a properly normalized Nearest Class Mean (NCM) rule acts as a competitive baseline. More recently, \citet{wang2024hardtobeat} showed that a training-free Linear Discriminant Analysis (LDA) classifier yields exceptionally strong performance when ensembled with zero-shot CLIP. Several studies \citep{goswami2026cross, zhang2026multi} have explored leveraging the text subspace to enhance CLIP's performance in the training-free few-shot setting. However, these methods typically rely on ad-hoc heuristics to align visual prototypes within this subspace. Specifically, TAMP \citep{goswami2026cross} projects few-shot image prototypes onto the text embedding space and linearly mixes the result with the text prototype via a scalar weight $\lambda$, which is selected on a held-out validation set. Alternatively, \citet{zhang2026multi} split the image embeddings into components within the text embedding space and its orthogonal complement, subsequently reweighting them via a tuned hyperparameter $\alpha$. Further discussion of related literature is provided in Appendix~\ref{app:related_work}.

\paragraph{Contributions} Our primary contributions in this paper are summarized as follows:

\begin{enumerate}
    \item We empirically demonstrate that while original zero-shot text embeddings is suboptimal for direct classification, the linear subspace they span is highly discriminative. Specifically, we show that Nearest Class Mean (NCM) classification, which uses image prototypes estimated strictly within this text-spanned subspace, \textcolor{blue}{often} achieves comparable or superior performance to full-space estimation while being significantly more sample-efficient.
    
    \item Building on these insights, we introduce DEFTS: a \textbf{D}ata-\textbf{E}fficient, training-free \textbf{F}ew-shot learning method in the \textbf{T}ext-embedding \textbf{S}ubspace. Importantly, we notice that because any point in this space is a linear combination of the text embeddings, prototype estimation reduces to inferring a low-dimensional set of coefficients. DEFTS then determines these coefficients via a closed-form Bayesian formulation, utilizing a prior naturally anchored to the original zero-shot text coordinates.
    
    \item We conduct extensive evaluations across 12 image classification benchmarks and three model backbones. \textcolor{blue}{The results demonstrate that DEFTS a highly effective complement to prior work: with model backbone ViT-L/14, ensembling DEFTS with existing methods yields a significantly positive improvement in 366 of the 480 dataset-level results (12 datasets, 8 methods, 5 numbers of shots). Only 19 results are negative, and only 5 of these are statistically significant (90\% confidence interval)}.
\end{enumerate}

\section{Method}
\label{sec:method}

In this section, we first review CLIP for zero-shot learning. Then we present a
series of observations demonstrating that the text embedding subspace is highly discriminative and
data efficient. Based on these insights, we provide details of our proposed training-free method DEFTS.

\begin{figure}[t]
\centering
\includegraphics[width=\linewidth]{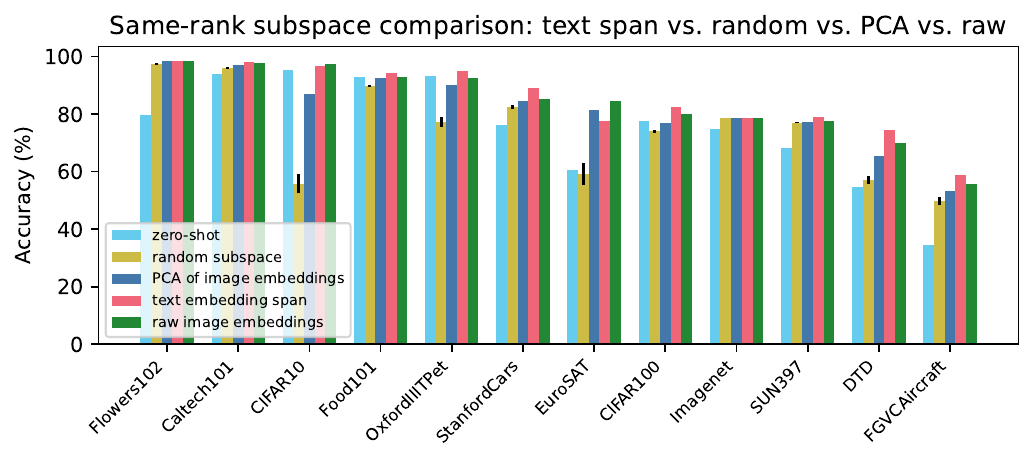}
\caption{Nearest Class Mean (NCM) accuracy across 12 datasets when projecting image embeddings into the text-spanned subspace. This is compared against zero-shot CLIP, the full un-projected image embedding space, a random subspace and a PCA subspace of the image embeddings both with rank equivalent to the text subspace. Results for the random subspace are averaged across 10 random seeds, with error bars indicating one standard error. \rt{The number of classes for each dataset and the number of samples for estimating the embeddings in the full and text-spanned subspace are included in Table~\ref{tab:dataset_statistics}.}}
\label{fig:subspace_comparison}
\end{figure}

\subsection{CLIP for zero-shot classfication}
\label{sec:clip}

In the zero-shot setting, CLIP performs classification by comparing an image embedding against a set of class-specific text embeddings, assigning the image to the class that yields the highest cosine similarity. Specifically, let $I$ denote an input image and $T_{1}, \dots, T_{K}$ denote the synthesized text prompts (e.g., ``a photo of a {class}'') for each of the $K$ classes. CLIP extracts the representations using a vision encoder $F_{v}$ (e.g., a Vision Transformer or ResNet) and a text encoder $F_{t}$ (typically a Transformer). Assuming the extracted embeddings are $L_{2}$-normalized, the predicted class $\hat{y}$ is determined by maximizing their inner product:
\begin{align} \label{eqn:clip}
\hat{y} = \arg\max_{i \in \{1, \dots, K\}} F_{v}(I)^\top F_{t}(T_i).
\end{align}

\subsection{Text Embedding subspace}
\label{sec:textsubspace}

As shown above, the rule for deciding the class in CLIP is straightforward.
However, it is less immediately obvious that the prediction results can
only depend on the image embedding through its projection onto the subspace spanned by the $K$
class text embeddings. \rt{To formalize this, let $e_i = F_{t}(T_i)$ denote the text embedding for class $i$, and $E = [e_1, \dots, e_K] \in \mathbb{R}^{d\times K}$ be the matrix of all class embeddings. We define $\mathcal{S} = \operatorname{span}(e_1,\dots,e_K)$ and the orthogonal projector onto $\mathcal{S}$ as $P_{\mathcal{S}} = E (E^\top E)^{+} E^\top$, where $^{+}$ indicates the pseudo-inverse. We can then decompose the visual embedding $F_v(I)$ as follows:
\begin{align*}
F_{v}(I) =  P_{\mathcal{S}} F_{v}(I) + (I - P_{\mathcal{S}}) F_{v}(I).
\end{align*}
As an orthogonal projection matrix, $P_{\mathcal{S}}$ is symmetric and idempotent. Because $e_i$ lies within the span of $\mathcal{S}$, it is invariant under this projection ($P_{\mathcal{S}} e_i = e_i$), which gives $(I - P_{\mathcal{S}}) e_i = 0$. Using the symmetry of $(I - P_{\mathcal{S}})$, it follows that $F_{v}(I)^\top (I - P_{\mathcal{S}}) e_i = 0$. Therefore, the dot product between the visual and text embeddings simplifies to:
\begin{align*}
F_{v}(I)^\top e_i = (P_{\mathcal{S}} F_{v}(I))^\top e_i = F_{v}(I)^\top P_{\mathcal{S}} e_i.
\end{align*}}\rt{A known reason why the raw text embeddings $e_i$ are poor prototypes is the modality gap \citep{liang2022mindthegap}: image and text embeddings occupy different regions of the embedding sphere, so the images of a class are not centered at the text embedding of that class. Thus,
prior work typically focus on looking for better alternatives to the text embedding $e_i$ as a class prototype by learning better representations \citep{zhou2022learning, goswami2026cross}, often ignoring the text embeddings entirely \citep{wang2024hardtobeat}.}

\textcolor{blue}{In this section, we ask whether the text subspace itself, rather than the text embedding vectors, is the reason that zero-shot classification is sub-optimal.} Informally, if $\mathcal{S}$ is an inherently poor subspace, projecting the image embeddings 
onto it will inevitably discard critical discriminative information. In this case, 
classification accuracy would be strictly bottlenecked by the projection. Conversely, if 
$\mathcal{S}$ is highly discriminative and preserves this information, then the 
raw text embeddings themselves are the true bottleneck limiting performance, not 
the subspace they span.

\begin{figure}[t]
\centering
\includegraphics[width=1\linewidth]{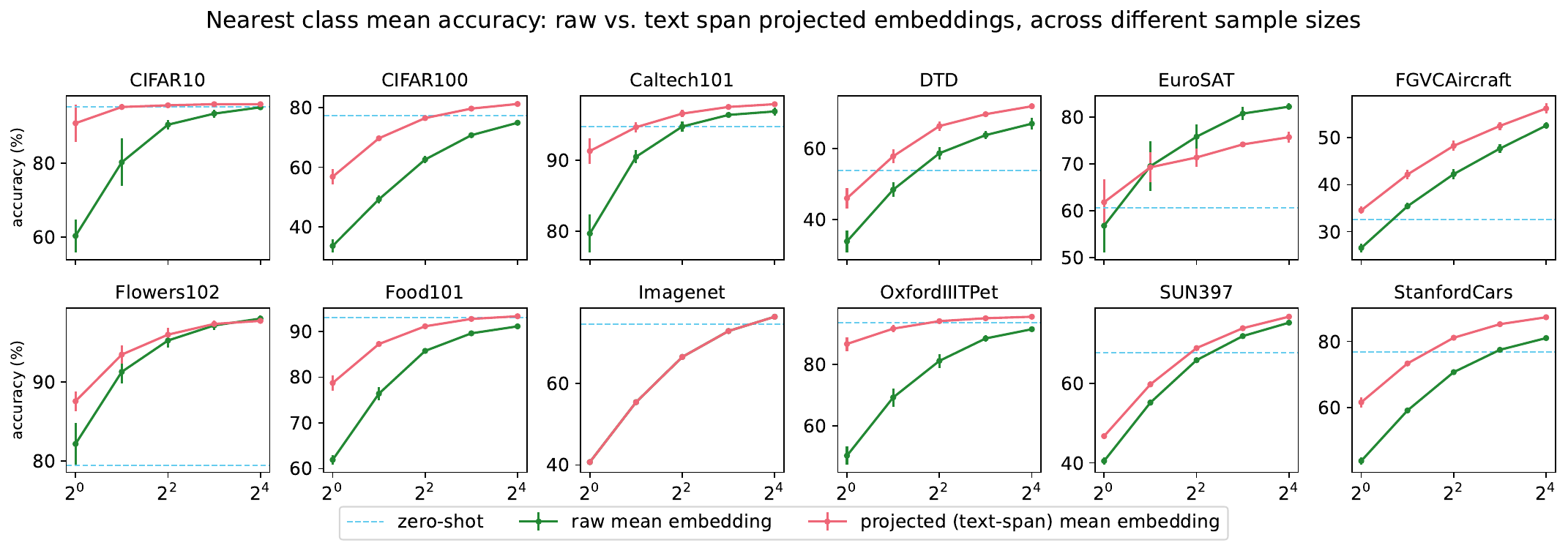}
\caption{Nearest Class Mean (NCM) accuracy across 12 datasets, 
comparing image prototypes estimated within the text subspace
against estimations in the un-projected embedding space. 
Accuracies are evaluated across varying $k$-shots, 
with all results averaged over 10 random seeds. \rt{The number of classes for each dataset and the number of samples for estimating the embeddings in the full and text-spanned subspace are included in Appendix~\ref{tab:dataset_statistics}.}}
\label{fig:data-efficiency}
\end{figure}

In this section, we present a series of observations for 
investigating the the text embedding subspace. All experiments 
conducted in this section employ CLIP ViT-L/14 as model backbone,
and utilize the hand-crafted prompt templates introduced by CoOp \citep{zhou2022learning}
as detailed in Appendix~\ref{app:prompt}. The experiments details
for the observations are provided in Appendix~\ref{app:experimentsobs}.

\paragraph{Observation 1: The text-spanned subspace is highly discriminative and not the bottleneck.}
In our first empirical observation, we investigate the discriminative power of the linear subspace spanned by the text embeddings (as described in Section~\ref{sec:bayes-pca} but without centering first). It is well established that replacing text embeddings with the class means of the image embeddings, effectively applying a Nearest Class Mean (NCM) classifier \citep{wang2019simpleshot}, significantly improves classification performance. To investigate the discriminative power of the text subspace, we train 
an NCM classifier strictly within this subspace, and subsequently evaluate its accuracy.

As shown in Figure~\ref{fig:subspace_comparison}, the accuracy of this projected NCM classifier is significantly higher than that of standard zero-shot CLIP. This suggests that the text subspace is \textbf{not} the bottleneck for classification accuracy. For comparison, we also evaluate 
the full un-projected image embedding space, projections onto a random subspace and a PCA subspace of the image embeddings both with rank equivalent to the text subspace.
Surprisingly, the text embedding subspace consistently outperforms the PCA subspace and, in several datasets, such as Food101, surpasses the full, un-projected image embeddings.

\paragraph{Observation 2: Class prototype estimation within the subspace is highly data-efficient.}
Following Observation 1, we further investigate the accuracy of the NCM classifier when 
only a small number of training samples per class are available to estimate the image prototypes. 
We compare the results of estimating these prototypes strictly within the text-spanned subspace against estimating them directly in the full embedding space. As shown in Figure~\ref{fig:data-efficiency}, subspace estimation is significantly more data-efficient \textcolor{blue}{for 10 of the 12 datasets considered}. It outperforms direct estimation by a large margin when the number of sample sizes are small with only two exceptions: ImageNet and EuroSAT. For ImageNet, the large number of classes ($K=1000$) causes the text embedding subspace to span the entire feature space, so the projection is equivalent to operating in the full space. \textcolor{blue}{To further investigate whether the number of classes $K$ impacts Observation 2. We provide additional experiments in Appendix~\ref{app:kd} analyzing the correlation between the ratio $K/d$ and the advantage of estimtion in text subspace over full space.} On the other hand, EuroSAT is a well-documented outlier to the CLIP pre-training distribution \citep{li2025recent, radford2021learning}, which limits the performance of the text subspace.

\begin{figure}[t]
\centering
\includegraphics[width=\linewidth]{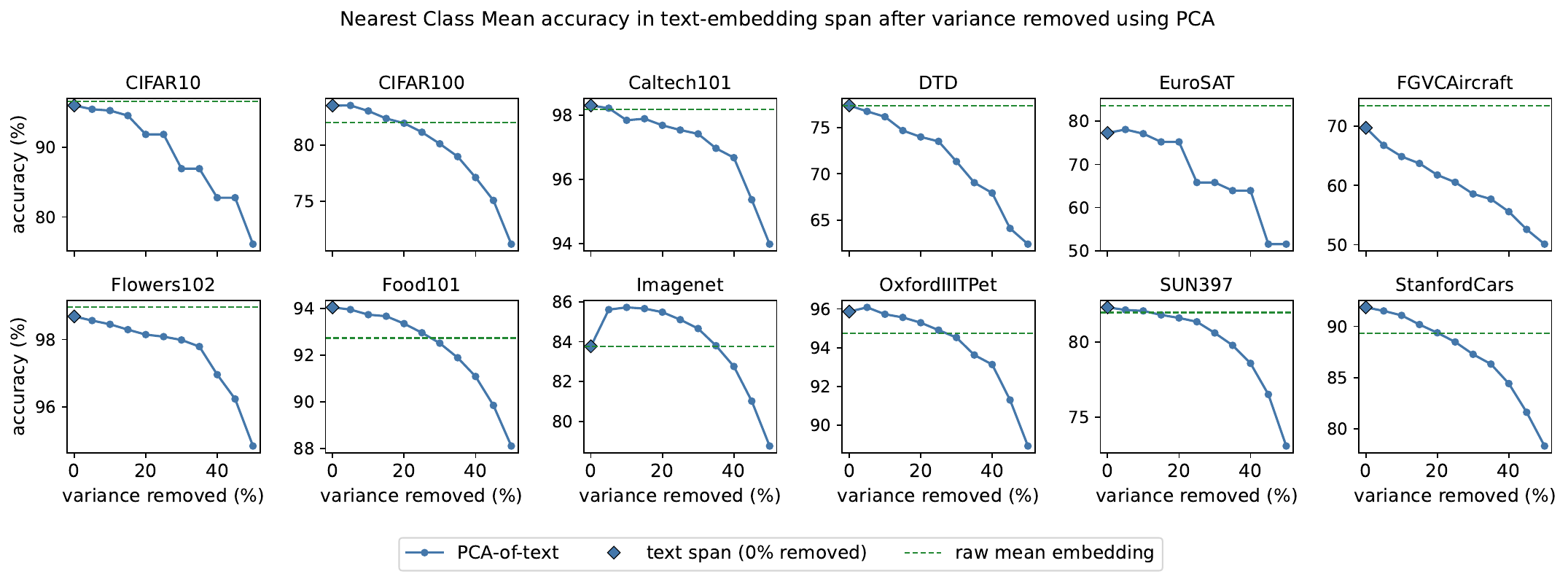}
\caption{Nearest Class Mean (NCM) classification accuracy across 12 datasets, 
evaluated within the text embedding subspace after PCA-based variance reduction.}
\label{fig:pca}
\end{figure}

\paragraph{\rt{Observation 3: The text-spanned subspace is tight, so only a mild reduction is useful in practice.}} In this observation, we investigate whether reducing the dimensionality of the text embedding subspace via PCA degrades its discriminative capacity. We evaluate this by projecting all training samples first and then measuring the NCM accuracy within the reduced subspace. As shown in Figure~\ref{fig:pca}, even a minor removal of variance, e.g.\ 5\%, hurts the discriminative power of the subspace, so no direction of the subspace is redundant. The only exception to this trend is ImageNet, whose 1000 text embeddings span the entire feature space. Note, however, that the image embeddings in Figure~\ref{fig:pca} are estimated from all available training samples. When only a few samples are available, as in this paper, discarding a minimal amount of variance (e.g., 5\%) improves the accuracy, whereas a large reduction (e.g., 50\%) strongly degrades it (Appendix~\ref{app:pca-variance}).

This section investigates the text embedding subspace through a series of 
observations. We show that this linear subspace is highly 
discriminative and in-subspace prototype estimation is \rt{often} significantly 
more data-efficient than directly estimating image embeddings in the full 
feature space. While all experiments in this section apply the hand-crafted 
prompt templates introduced by CoOp \citep{zhou2022learning}, we present evidence in 
Appendix~\ref{app:template} showing that, surprisingly, the discriminative power of 
the text-spanned subspace does not seem to depend on the quality of the prompt template.

\subsection{DEFTS}
\label{sec:defts}
\rt{Section~\ref{sec:textsubspace} demonstrates empirically that the text embedding space is highly discriminative and data efficient. In this section, we introduce DEFTS, which performs NCM classification with image prototypes estimated strictly within the text-spanned linear subspace.

This estimation problem has two features: first, in the few-shot setting, the sample mean of the image coordinates computed directly from the $k$ images per class is a noisy estimate of the true prototype. Second, the text embedding of a class already provides a good guess of its image prototype after projection, because zero-shot CLIP classifies well. Therefore, instead of estimating the sample mean from scratch, we combine the two features using a Bayesian treatment. Specifically,
we place a Gaussian prior centered at the projected text embedding and model the image coordinates with a Gaussian likelihood. This gives us a closed-form solution: the posterior mean is a weighted average of the image mean and the text embedding, where the weight is estimated from the data directly.

Further, we consider two designs to keep the estimation problem more reliable in the few-shot regime. First, projecting the image embeddings onto the text subspace reduces the dimension from $d$ to $r \ll d$, so that only $r$ coefficients per class have to be inferred and the covariance is an $r \times r$ matrix. Second, the prior uses a single variance $\tau^2$, which can be estimated reliably from the $K$ deviations between the image means and the text prior, and the likelihood covariance is shared across classes, which pools all $K(k-1)$ degrees of freedom. In short, the subspace projection is designed to addresses the dimensionality of the problem, and the Bayesian update is for reducing the noise from the few samples. In Section~\ref{sec:ablation}, we validate the two design choices by showing that each of the two improves the accuracy of DEFTS on its own.}

\paragraph{Text Embedding Subspace Projection.}\label{sec:bayes-pca}
We first describe how we apply PCA to further reduce the dimensionality of the text embedding subspace. 
Let $e_1,\dots,e_K\in\mathbb{R}^d$ denote the CLIP text embeddings for the $K$ classes 
(generated from a fixed prompt template), and let $\bar{e} = \frac{1}{K}\sum_{c=1}^K e_c$ be their mean. We construct the centered scatter matrix:
\begin{align*}
S = \sum_{c=1}^K (e_c-\bar{e})(e_c-\bar{e})^\top \in \mathbb{R}^{d\times d}.
\end{align*}
Then we eigendecompose $S$ and retain the top $r$ eigenvectors to form the basis matrix $Q\in\mathbb{R}^{d\times r}$, where $r$ is chosen to capture at least 95\% of the total variance. Specifically, $r$ is the smallest integer satisfying $\sum_{i=1}^r \lambda_i / \sum_{i=1}^K \lambda_i \ge 0.95$. As demonstrated by Observation 3 in Section~\ref{sec:textsubspace} and the empirical results in Appendix~\ref{app:pca-variance}, this 95\% threshold provides practical benefits and improves 
classification accuracy when only a small number of training samples are available, whereas further reduction in variance decreases the discriminative power of the subspace. Note that it is also possible to project the image embeddings onto the original, uncentered text embeddings, as we have done 
in Section~\ref{sec:textsubspace}. In Appendix~\ref{app:pcacenter}, we compare these centered and uncentered projection spaces, and show that centering the text embeddings prior to projection yields improved classification accuracy.

In the following, we project all embeddings into this $r$-dimensional subspace. For each class $c$, the text embedding coordinate is $e_c' = Q^\top e_c \in \mathbb{R}^r$, and the $k$ support images $x_{c,i}$ yield coordinates $b_{c,i} = Q^\top x_{c,i} \in \mathbb{R}^r$. 
All subsequent Bayesian estimation operates entirely within this reduced coordinate space.

\paragraph{Bayesian Formulation: Prior and Likelihood}
\label{sec:bayes-model}

Let $\theta_c \in \mathbb{R}^r$ represent the unknown true prototype for class $c$ in the 
projected subspace. We place a Gaussian prior on $\theta_c$, centered exactly 
at the projected text embedding $e_c'$:
\begin{align*}
    \theta_c \sim \mathcal{N}(e_c', \tau^2 I_r).
\end{align*}
This prior is motivated by utilizing CLIP's zero-shot capabilities. 
As demonstrated in Section~\ref{sec:clip}, CLIP's classification rule depends on the 
image embedding strictly through its projection onto the text-spanned subspace. 
Because zero-shot classification is generally effective, the true image prototype 
naturally align with, and therefore exhibit a small angle relative to its 
corresponding text embedding in this space. Consequently, the projected text 
embedding $e_c'$ serves as a highly informative prior for the mean of image embeddings.

The $k$ observed support image coordinates are modeled as independent and identically distributed (i.i.d.) draws from a Gaussian likelihood parameterized by 
$\theta_c$:
\begin{align*}
b_{c,i} \mid \theta_c \sim \mathcal{N}(\theta_c, \Sigma), \qquad i=1,\dots,k.    
\end{align*}
Note that the covariance matrix $\Sigma$ does not depend on the class index $c$ 
and is instead shared across all classes. We adopt this shared covariance structure 
to maximize data efficiency in the few-shot settings, as demonstrated in Section~\ref{sec:pooled}.

\paragraph{Empirical Bayes Hyperparameter Estimation}
\label{sec:bayes-hyperparams}
To estimate the hyperparameters of the prior and likelihood, we employ the Empirical Bayes method. 
Let $\bar{b}_c = \frac{1}{k}\sum_{i=1}^k b_{c,i}$ denote the sample mean of 
the projected images for class $c$, and let $\delta_c = \bar{b}_c - e_c'$ 
be its deviation from the text prior. For $k \ge 2$, we first estimate the 
observation noise by computing the pooled sample variance across all $K$ classes:
\begin{align}\label{eqn:sigma}
\hat{\sigma}^2 = \frac{1}{r K(k-1)} \sum_{c=1}^K \sum_{i=1}^k \Vert b_{c,i} - \bar{b}_c \Vert^2.    
\end{align}
Then we estimate $\tau^2$ as:
\begin{align} \label{eqn:tau}
    \hat{\tau}^2 = \frac{1}{r} \overline{\Vert\delta\Vert^2} - \frac{\hat{\sigma}^2}{k},
\end{align}
where $\frac{1}{r} \overline{\Vert\delta\Vert^2} = \frac{1}{rK}\sum_{c=1}^K\left(\left\Vert \delta_c \right\Vert^2\right)$ is the averaged deviation of the image mean from the text prior to 
approximate $\frac{1}{r}\mathbb{E}\left[\Vert\bar{b}_c - e_c\Vert^2\right]$.
Appendix~\ref{app:hyperparameter_partone} shows the derivation for the estimator.

In the one-shot setting ($k=1$), the within-class variance $\hat{\sigma}^2$ 
cannot be computed from the data. Consequently, $\sigma^2$ and $\tau^2$ become 
unidentifiable, as only their sum, $\frac{1}{r}\overline{\Vert\delta\Vert^2} = \tau^2 + \sigma^2$, is observed. To resolve this problem, we partition the total observed spread using empirical results when $k > 1$ as shown in Appendix~\ref{app:contribution}. 
These results indicate that $\hat{\sigma}^2$ often accounts for roughly 10\% of the averaged empirical deviation $\frac{1}{r} \overline{\Vert\delta\Vert^2}$. Thus, we
set $\hat{\sigma}^2 = f\frac{1}{r} \overline{\Vert\delta\Vert^2}$ and $\hat{\tau}^2 = (1-f)\frac{1}{r} \overline{\Vert\delta\Vert^2}$ for $f = 0.1$ throughout the whole paper.

For the likelihood covariance, we apply a ridge-type shrinkage estimator analogous to the formulation in \citet{wang2024hardtobeat}, adapted here for the $r$-dimensional subspace. 
Let $\hat{\Sigma} = \frac{1}{K(k-1)} \sum_{c=1}^K \sum_{i=1}^k (b_{c,i} - \bar{b}_c)(b_{c,i} - \bar{b}_c)^\top$ be the pooled empirical sample covariance. The regularized precision matrix is defined globally for all classes as:
\begin{align} \label{eqn:covariance}
\Sigma^{-1} = \beta (K(k-1) + 2r + 1) \left[ K(k-1)\hat{\Sigma} + \operatorname{tr}(\hat{\Sigma})I_r \right]^{-1}, \qquad k \ge 2,
\end{align}
where $\beta \in (0, 1]$ is a covariance inflation parameter. Because empirical sample covariances derived from extremely small support sets (e.g. $k = 2$) could be highly unstable as shown in Appendix~\ref{app:covariance}, the implicit factor $\frac{1}{\beta}$ explicitly inflates the estimated variance to prevent the Bayesian update from becoming overconfident. We investigate the impact of this hyperparameter in detail in Appendix~\ref{app:beta}. Empirically, we observe that a small $\beta$ (e.g., $0.1$ or $0.2$) yields better performance than $\beta = 1$, particularly in low-shot regimes such as $k = 2$. Therefore, we fix $\beta = 0.1$ throughout the paper. \rt{In Appendix~\ref{app:persens}, we conduct a sensitivity analysis for this hyperparameter alongside the PCA variance threshold, demonstrating their robustness across different datasets.} For the 1-shot setting ($k=1$), we default to $\Sigma^{-1} = \beta \hat{\sigma}^{-2} I_r$. The full derivation for this estimate is provided in Appendix~\ref{app:hyperparameter_parttwo}.

\paragraph{Closed-Form Posterior Inference}
\label{sec:bayes-posterior}

Denote $\Omega_0 = \hat{\tau}^{-2} I_r$. Given the Gaussian prior and likelihood, the posterior distribution is also Gaussian, and its posterior mean is \citep{bishop2006pattern}:
\begin{align} \label{eqn:posterior}
    \hat{\theta}_c = (\Omega_0 + k\Sigma^{-1})^{-1} (\Omega_0 e_c' + k\Sigma^{-1} \bar{b}_c).
\end{align}
\rt{In other words, the posterior mean is a weighted average of the sample mean $\bar b_c$ and the prior mean $e'_c$. This is an empirical-Bayes shrinkage estimator in the sense of James--Stein~\citep{james1961estimation} and Efron--Morris~\citep{efron1975data}: the strength of the shrinkage is estimated from the data, through $\hat\sigma^2$ and $\hat\tau^2$, and not tuned by hand or require additional dataset for hyper-parameter tunning as TAMP \citep{goswami2026cross}.}

\rt{In Appendix~\ref{app:oneshot}, we show that when $k = 1$, the Equation~\eqref{eqn:posterior}
degrages to $\hat{\theta}_c  \approx 0.474 e_c' + 0.526 \bar{b}_c$ for $\beta = 0.1$ and $f = 0.1$. In other words, the weights of the image and text embedding does not depend on the data for DEFTS when $k = 1$. Therefore, the performances of DEFTS for $k=1$ depends solely on the centered PCA projection, rather than Bayesian updates.} 


The complete procedure for performing DEFTS is summarized in Algorithm~\ref{alg:defts} in Appendix~\ref{app:algorithm}. The results from DEFTS can be used directly for cosine-similarity classification, or ensembled with existing methods as described in Appendix~\ref{app:hypermethod}.

\section{Experiments}
\label{sec:experiments}

\subsection{DEFTS Complements Existing Training-Free Methods}
\label{sec:compare}

In this section, we conduct extensive experiments to evaluate the performance of DEFTS, and
existing training-free methods after complemented by DEFTS through ensembling.

\paragraph{Datasets.} Following prior relevant work \citep{wang2024hardtobeat, radford2021learning, zhou2022learning}, we evaluate DEFTS on 12 standard few-shot classification datasets. The datasets span generic object recognition (CIFAR-10/100~\citep{krizhevsky2009learning}, Caltech101~\citep{fei2004learning}, ImageNet~\citep{deng2009imagenet}), fine-grained recognition (StanfordCars~\citep{krause20133d}, FGVCAircraft~\citep{maji2013fine}, Flowers102~\citep{nilsback2008automated}, OxfordIIITPet~\citep{parkhi2012cats}, Food101~\citep{bossard2014food}), texture recognition (DTD~\citep{cimpoi2014describing}), scene recognition (SUN397~\citep{xiao2010sun}), and remote-sensing imagery (EuroSAT~\citep{helber2019eurosat}). In Appendix~\ref{app:dataset}, we provide more information about these datasets.

\begin{table*}[htbp]
\centering
\caption{Number of shots $k=2$. Classification accuracy (\%) and performance improvements across datasets. The values in parentheses denote the absolute performance change after ensembling with DEFTS and the $90\%$ confidence interval.}
\label{tab:ensemble_improvements_datasets_1}

\newcommand{\gain}[3]{\begin{tabular}{@{}c@{}}#1 \\[-1ex] \textcolor{green!60!black}{\tiny (+#2 $\pm$#3)}\end{tabular}}
\newcommand{\loss}[3]{\begin{tabular}{@{}c@{}}#1 \\[-1ex] \textcolor{red!80!black}{\tiny (-#2 $\pm$#3)}\end{tabular}}

\renewcommand{\arraystretch}{1.0} 
\setlength{\tabcolsep}{4.5pt}

\resizebox{\textwidth}{!}{
\scriptsize 
\begin{tabular}{@{}l ccccccccc@{}}
\toprule
\textbf{Dataset} & \textbf{ZS} & \textbf{SimpleShot} & \textbf{Linear Probe} & \textbf{Tip-Adapter} & \textbf{APE} & \textbf{TIP-X} & \textbf{TAMP+ZS} & \textbf{LDA+ZS} & \textbf{DEFTS} \\
\midrule
C10 & 
\gain{94.89}{0.38}{0.18} & \gain{79.44}{15.84}{3.08} & \gain{81.42}{13.90}{2.77} & \gain{94.85}{0.24}{0.24} & \gain{94.34}{0.45}{0.61} & \gain{94.48}{0.54}{0.19} & \gain{94.80}{0.39}{0.16} & \gain{94.80}{0.41}{0.12} & 95.29 \\
\addlinespace[2pt]
C100 & 
\gain{77.27}{0.29}{0.30} & \gain{48.35}{28.50}{0.67} & \gain{53.39}{23.38}{2.10} & \gain{77.53}{0.00}{0.00} & \gain{77.65}{0.00}{0.00} & \gain{68.43}{2.32}{1.02} & \gain{77.38}{0.16}{0.18} & \gain{77.27}{0.29}{0.30} & 76.85 \\
\addlinespace[2pt]
Cal101 & 
\gain{94.62}{0.53}{0.31} & \gain{91.13}{4.47}{0.48} & \gain{91.62}{3.70}{0.72} & \gain{94.64}{0.00}{0.00} & \gain{94.83}{0.00}{0.00} & \gain{94.59}{0.53}{0.12} & \gain{94.95}{0.41}{0.28} & \gain{94.62}{0.53}{0.31} & 95.58 \\
\addlinespace[2pt]
DTD & 
\gain{53.03}{4.98}{1.87} & \gain{48.05}{14.29}{1.38} & \gain{48.95}{14.47}{2.48} & \gain{57.09}{1.70}{1.64} & \gain{56.33}{0.31}{0.57} & \gain{56.20}{2.18}{0.91} & \gain{56.82}{1.28}{1.23} & \gain{53.03}{4.98}{1.87} & 62.37 \\
\addlinespace[2pt]
Euro & 
\gain{61.17}{5.66}{2.26} & \gain{67.60}{5.18}{2.74} & \gain{64.66}{8.25}{4.37} & \gain{66.87}{1.39}{0.94} & \gain{68.90}{2.06}{2.67} & \gain{66.81}{1.77}{0.54} & \gain{61.07}{6.26}{2.24} & \gain{67.54}{4.27}{3.14} & 68.67 \\
\addlinespace[2pt]
Air. & 
\gain{32.18}{0.25}{0.47} & \gain{34.62}{8.55}{1.71} & \gain{33.55}{10.36}{1.46} & \gain{34.76}{0.00}{0.00} & \gain{37.49}{0.74}{1.35} & \gain{43.29}{1.90}{0.37} & \gain{35.56}{0.15}{0.32} & \gain{32.18}{0.25}{0.47} & 44.57 \\
\addlinespace[2pt]
Flow. & 
\gain{78.40}{15.16}{0.94} & \gain{92.75}{2.85}{0.66} & \gain{91.44}{3.91}{1.22} & \gain{88.02}{5.71}{1.77} & \gain{93.00}{1.19}{0.59} & \gain{92.84}{1.66}{0.23} & \gain{91.22}{1.15}{0.57} & \gain{92.81}{1.64}{1.12} & 95.14 \\
\addlinespace[2pt]
Food & 
\gain{93.11}{0.23}{0.11} & \gain{75.70}{17.06}{0.56} & \gain{77.23}{15.33}{1.07} & \gain{93.19}{0.00}{0.00} & \gain{93.19}{0.00}{0.00} & \gain{87.53}{2.34}{0.42} & \gain{93.17}{0.21}{0.10} & \gain{93.11}{0.23}{0.11} & 92.76 \\
\addlinespace[2pt]
INet & 
\gain{74.35}{0.00}{0.00} & \gain{54.78}{20.90}{0.27} & \gain{57.89}{17.77}{0.23} & \gain{74.40}{0.00}{0.00} & \gain{74.50}{0.00}{0.00} & \gain{64.45}{0.00}{0.00} & \gain{75.68}{0.00}{0.00} & \gain{74.35}{0.00}{0.00} & 75.68 \\
\addlinespace[2pt]
Pet & 
\gain{92.82}{0.88}{0.30} & \gain{66.96}{26.31}{1.72} & \gain{70.14}{23.02}{1.69} & \gain{92.87}{0.12}{0.23} & \gain{93.01}{0.13}{0.24} & \gain{93.42}{0.35}{0.45} & \gain{92.85}{1.17}{0.25} & \gain{92.82}{0.88}{0.30} & 93.27 \\
\addlinespace[2pt]
SUN & 
\gain{67.35}{0.00}{0.00} & \gain{55.13}{17.03}{0.38} & \gain{56.14}{16.02}{1.04} & \gain{67.92}{0.00}{0.00} & \gain{68.05}{0.00}{0.00} & \gain{63.45}{0.00}{0.00} & \gain{69.40}{0.00}{0.00} & \gain{67.35}{0.00}{0.00} & 72.16 \\
\addlinespace[2pt]
Cars & 
\gain{76.61}{0.00}{0.00} & \gain{59.02}{22.59}{0.79} & \gain{62.43}{18.77}{1.27} & \gain{76.72}{0.00}{0.00} & \gain{76.82}{0.00}{0.00} & \gain{76.83}{1.62}{0.65} & \gain{76.89}{0.04}{0.05} & \gain{76.61}{0.00}{0.00} & 81.64 \\
\midrule
AVG & 
\gain{\textbf{74.65}}{2.36}{0.69} & \gain{\textbf{64.46}}{15.30}{1.30} & \gain{\textbf{65.74}}{14.07}{1.07} & \gain{\textbf{76.57}}{0.76}{0.31} & \gain{\textbf{77.34}}{0.41}{0.25} & \gain{\textbf{75.19}}{1.27}{0.18} & \gain{\textbf{76.65}}{0.94}{0.32} & \gain{\textbf{76.37}}{1.12}{0.37} & \textbf{79.50} \\
\bottomrule
\end{tabular}
}
\end{table*}

\paragraph{Existing methods.} We consider completing the following training-free 
few-shot classification methods using DEFTS:  \textbf{Linear Probe}~\citep{radford2021learning}:
a linear classier on top of image embeddings;
\textbf{SimpleShot}~\citep{wang2019simpleshot}, 
a nearest class mean classifier; \textbf{Tip-Adapter}~\citep{zhang2022tipadapter}, 
which caches the $k$-shot features as keys and one-hot labels as values to compute 
a weighted cache vote added to the zero-shot logits; \textbf{TIP-X}~\citep{udandarao2023susx}, 
which extends Tip-Adapter with a KL-divergence affinity term between the support and 
test images' text-class distributions; 
\textbf{APE}~\citep{zhu2023not}, which refines Tip-Adapter by selecting class-discriminative 
channels from text embeddings and reweighting cache votes; \textbf{LDA}~\citep{wang2024hardtobeat}, a Gaussian discriminant classifier utilizing a
single empirical Bayes shrunk covariance pooled across all classes; and \textbf{TAMP}~\citep{goswami2026cross}, 
which mixes image and text prototypes in an un-truncated text embeddings subspace. 
As the standalone LDA classifier typically yields sub-optimal performance, we ensemble it with 
zero-shot CLIP, as suggested by \citet{wang2024hardtobeat}. Similarly, we evaluate the performances of TAMP after ensembling with both zero-shot CLIP.

\paragraph{Implementation Details.} All embeddings are extracted using a frozen CLIP ViT-L/14~\citep{radford2021learning} and prompt templates from CoOp~\citep{zhou2022learning} as detailed in Appendix~\ref{app:prompt}. We evaluate $k\in\{1,2,4,8,16\}$ support images per class, with results averaged over 10 random seeds. Ensembling methods require choosing weights for each method, e.g. the weight between LDA and zero-shot (ZS). In this section, all methods strictly adhere to the $k$-shot budget without utilizing external held-out validation. Thus, when $k = 1$, we apply fixed weights for each method. When $k \ge 2$, we search optimal hyperparameters for each method using leave-one-out cross-validation. The details for choosing and tuning these hyperparameters is reported in Appendix~\ref{app:hyper}.

\begin{table*}[htbp]
\centering
\caption{Average classification accuracy (\%) and performance improvements across 12 datasets for different numbers of shots $k$. The values in parentheses denote the absolute performance change after ensembling with DEFTS and the $90\%$ confidence interval. Positive gains are highlighted in green. The best baseline accuracy for each $k$-shot setting (before adding DEFTS) is highlighted in \textbf{bold}.}
\label{tab:ensemble_improvements_k_shots}

\newcommand{\gain}[3]{\begin{tabular}{@{}c@{}}#1 \\[-1ex] \textcolor{green!60!black}{\tiny (+#2 $\pm$#3)}\end{tabular}}
\newcommand{\loss}[3]{\begin{tabular}{@{}c@{}}#1 \\[-1ex] \textcolor{red!80!black}{\tiny (-#2 $\pm$#3)}\end{tabular}}

\renewcommand{\arraystretch}{1.0} 
\setlength{\tabcolsep}{4.5pt}

\resizebox{\textwidth}{!}{
\scriptsize 
\begin{tabular}{@{}l ccccccccc@{}}
\toprule
\textbf{$k$-shot} & \textbf{ZS} & \textbf{SimpleShot} & \textbf{Linear Probe} & \textbf{Tip-Adapter} & \textbf{APE} & \textbf{TIP-X} & \textbf{TAMP+ZS} & \textbf{LDA+ZS} & \textbf{DEFTS alone} \\
\midrule
$k=1$ & 
\gain{74.65}{3.03}{0.46} & \gain{50.88}{18.37}{1.25} & \gain{55.27}{16.43}{1.15} & \gain{75.05}{2.69}{0.43} & \gain{75.79}{2.07}{0.38} & \gain{75.29}{2.48}{0.41} & \gain{76.21}{1.02}{0.21} & \gain{74.69}{3.00}{0.45} & \textbf{77.12} \\
\addlinespace[2pt]
$k=2$ & 
\gain{74.65}{2.36}{0.69} & \gain{64.46}{15.30}{1.30} & \gain{65.74}{14.07}{1.07} & \gain{76.57}{0.76}{0.31} & \gain{77.34}{0.41}{0.25} & \gain{75.19}{1.27}{0.18} & \gain{76.65}{0.94}{0.32} & \gain{76.37}{1.12}{0.37} & \textbf{79.50} \\
\addlinespace[2pt]
$k=4$ & 
\gain{74.65}{6.33}{0.91} & \gain{73.83}{8.48}{0.73} & \gain{75.84}{6.35}{0.66} & \gain{79.05}{2.16}{0.45} & \gain{79.78}{1.64}{0.36} & \gain{76.93}{0.52}{0.13} & \gain{78.80}{2.20}{0.46} & \gain{\textbf{82.12}}{0.68}{0.26} & 81.66 \\
\addlinespace[2pt]
$k=8$ & 
\gain{74.65}{8.90}{1.10} & \gain{78.80}{5.54}{0.47} & \gain{81.44}{2.61}{0.36} & \gain{81.37}{2.79}{0.35} & \gain{82.06}{2.29}{0.38} & \gain{78.84}{0.03}{0.06} & \gain{80.00}{3.30}{0.54} & \gain{\textbf{85.24}}{0.49}{0.13} & 83.51 \\
\addlinespace[2pt]
$k=16$ & 
\gain{74.65}{10.04}{1.20} & \gain{81.56}{4.00}{0.37} & \gain{85.12}{0.37}{0.27} & \gain{82.85}{2.69}{0.33} & \gain{83.84}{2.11}{0.35} & \gain{80.58}{0.43}{0.20} & \gain{80.89}{3.78}{0.59} & \gain{\textbf{86.70}}{0.70}{0.16} & 84.59 \\
\bottomrule
\end{tabular}
}
\end{table*}

\paragraph{Results} \rt{In Table~\ref{tab:ensemble_improvements_datasets_k1}, we present the classification results for the two-shot learning setting ($k = 2$). Hyperparameters and ensembling weights for each method were tuned following the procedure detailed in Appendix~\ref{app:hyper}. Note that a reported gain of $0.00 \pm 0.00$ indicates that the leave-one-out search assigned a weight of 0 to DEFTS across all 10 random seeds, effectively switching it off. 

The results demonstrate that ensembling with DEFTS improves the average accuracy of all evaluated methods across the 12 datasets. For weaker baselines, such as SimpleShot and Linear Probe, the performance gains yielded by DEFTS are substantial. Meanwhile, for strong baselines like LDA, we still observe statistically significant improvements. Notably, the standalone DEFTS framework achieves the highest overall average accuracy in this 2-shot case. In Appendix~\ref{app:detailedresults}, we provide detailed results analogous to Table~\ref{tab:ensemble_improvements_datasets_k1} for varying numbers of shots $k$.}

Table~\ref{tab:ensemble_improvements_k_shots} further evaluates the performance of each method across varying numbers of shots $k$. The results show that DEFTS consistently complements existing baselines. In extremely low-data regimes ($k \in \{1,2\}$), the standalone DEFTS framework achieves the highest overall accuracy while significantly improving the performance of all other methods upon ensembling. For larger support sets ($k \in \{4,8,16\}$), LDA+ZS emerges as the strongest initial baseline; crucially, ensembling it with DEFTS continues to yield statistically significant improvements. Finally, in Appendix~\ref{app:backbones}, we replicate these simulations using alternative model backbones, demonstrating that the performance gains provided by DEFTS are robust across different models.

\subsection{Ablation Study}
\label{sec:ablation}


\subsubsection{Estimation in subspace}

\begin{figure}[t]
    \begin{minipage}[b]{0.31\linewidth}
        \centering
        \includegraphics[width=\linewidth]{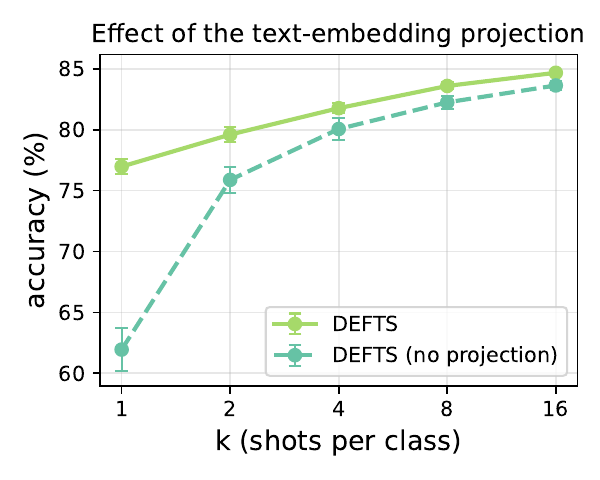}
        \caption{DEFTS' accuracy with and without projecting into the
        text subspaces.}
        \label{fig:ablation1}
    \end{minipage} \hfill
    \begin{minipage}[b]{0.31\linewidth}
        \centering
        \includegraphics[width=\linewidth]{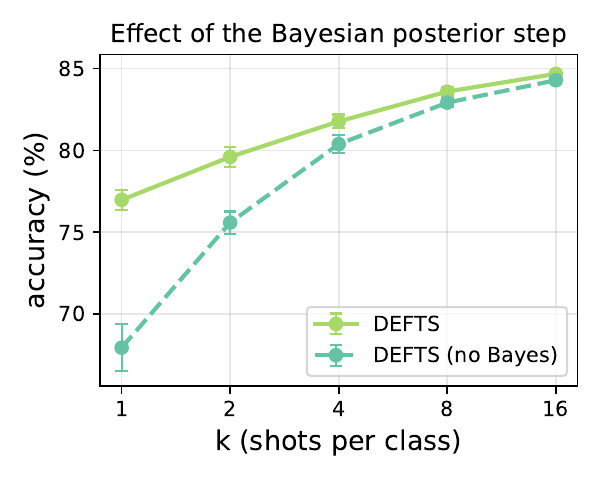}
        \caption{DEFTS' accuracy with and without applying the Bayesian
        estimation.}
        \label{fig:ablation2}
    \end{minipage} \hfill
    \begin{minipage}[b]{0.31\linewidth}
        \centering
        \includegraphics[width=\linewidth]{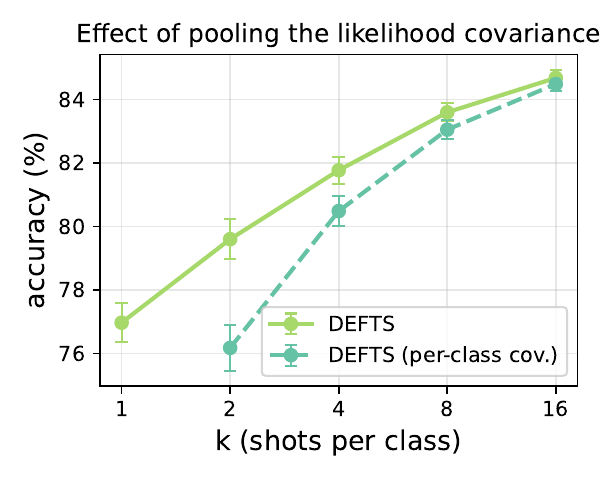}
        \caption{DEFTS' accuracy with and without applying the pooled covariance estimation
        .}
        \label{fig:ablation3}
    \end{minipage}
\end{figure}

DEFTS proposes projecting features into a low-rank, variance-reduced text embedding subspace prior to performing Bayesian estimation for the class-conditional image means. We first investigate the impact of this projection by comparing it against an ablation where Bayesian estimation is applied directly in the original, unprojected embedding space. The results Figure~\ref{fig:ablation1} demonstrate that projecting into the subspace yields a substantial advantage especially in the low-data regimes ($k \in \{1, 2\}$). As the number of shots $k$ increases, the performance of DEFTS with and without the subspace projection converges.

\subsubsection{Bayesian formulations}

We now investigate the contribution of the Bayesian estimation component within the DEFTS framework. In Figure~\ref{fig:ablation2}, we compare the DEFTS against a simple average of the image embeddings in the DEFTS' subspace. \rt{In Table~\ref{tab:abl_weight}, we further compare DEFTS against DEFTS (fixed weight 0.5), which mixes image and text embeddings evenly in DEFTS's centered 95\% subspace. Both comparisons confirm the effectiveness of the Bayesian component of DEFTS. According to Figure~\ref{fig:ablation2}, while DEFTS drastically outperforms the simple image average for $k < 8$, this gap narrows when $k = 16$. On the other hand, DEFTS' advantage over the fixed-weight variant increases with $k$ as shown inTable~\ref{tab:abl_weight}. This highlights the difficulty of choosing a reasonable mixing weight between image and text embeddings: 
relying on text embeddings is useful for small $k$, but forcing this reliance at 
larger $k$ degrades the performance. DEFTS naturally addresses this issue by adapting the 
mixing weight dynamically through the data.}

\begin{table}[!ht]
\centering
\caption{\rt{Classification accuracy of DEFTS and DEFTS (fixed weight 0.5), which mixes image and text embeddings in DEFTS's centered 95\% subspace evenly. Results are average over 10 random seeds.}}
\label{tab:abl_weight}
\rt{\small
\renewcommand{\arraystretch}{1.15}
\begin{tabular}{@{}l ccccc@{}}
\toprule
\textbf{Method} & 1 shot & 2 shots & 4 shots & 8 shots & 16 shots \\
\midrule
DEFTS (fixed weight 0.5) & 77.16 & 78.82 & 79.56 & 80.02 & 80.32 \\
DEFTS & 77.12 & 79.50 & 81.66 & 83.51 & 84.59 \\
\bottomrule
\end{tabular}}
\end{table}

\subsubsection{Pooled covariance estimation}
\label{sec:pooled}

As discussed in Section~\ref{sec:defts}, DEFTS relies on a pooled sample 
covariance matrix (Equation~\ref{eqn:covariance}) to estimate the 
distribution of the image coordinates. To justify this design choice, we conduct an ablation study comparing our pooled covariance strategy against independent per-class covariance estimation. The results are shown in Figure~\ref{fig:ablation3}. Note that in the one-shot setting ($k = 1$), computing a per-class covariance is mathematically intractable. When $k \in \{2, 4\}$, pooling the variance across all classes yields superior performance. When the support set size increases to $k = 8, 16$, 
the performances of per-class covariance estimation gets close to the pooled approach. 

\subsubsection{Comparison with closest baseline}

In previous ablation studies, we isolate each component of DEFTS 
and investigates their contributions for DEFTS. In this section,
we compare DEFTS directly with TAMP, serving as the closest baseline to DEFTS.
TAMP projects image embeddings into uncentered 
text embedding spaces and estimates the final representations 
via a weighted average of the mean image embeddings and 
text embeddings. This averaging weight is then tuned via leave-one-out cross-validation in this paper,
as described in Appendix~\ref{app:hyper}. The detailed theoretical comparison between DEFTS and TAMP can be seen in Appendix~\ref{sec:defts_vs_tamp}. As shown in Table~\ref{tab:tamp_vs_defts}, the differences between DEFTS and TAMP matters practically as well: DEFTS significantly outperforms TAMP across all evaluated $k$-shot settings.

\begin{table*}[t]
\centering
\caption{Comparison of average classification accuracy (\%) between TAMP and DEFTS across different numbers of shots $k$. The values in parentheses denote the $90\%$ confidence interval. Positive differences (DEFTS $-$ TAMP) are highlighted in green.}
\label{tab:tamp_vs_defts}

\newcommand{\val}[2]{#1\,{\tiny ($\pm$#2)}}
\newcommand{\gain}[2]{\textcolor{green!60!black}{#1\,{\tiny ($\pm$#2)}}}

\renewcommand{\arraystretch}{1.2} 
\setlength{\tabcolsep}{8pt} 

\scriptsize 
\begin{tabular}{@{}l ccccc@{}}
\toprule
\textbf{Method} & \textbf{$k=1$} & \textbf{$k=2$} & \textbf{$k=4$} & \textbf{$k=8$} & \textbf{$k=16$} \\
\midrule
TAMP alone & \val{74.47}{2.65} & \val{76.52}{2.72} & \val{78.81}{2.44} & \val{79.96}{2.33} & \val{80.89}{2.21} \\
DEFTS alone & \val{77.12}{2.49} & \val{79.50}{2.34} & \val{81.66}{2.16} & \val{83.51}{1.95} & \val{84.59}{1.83} \\
\midrule
DEFTS $-$ TAMP & \gain{+2.66}{0.35} & \gain{+2.97}{0.56} & \gain{+2.85}{0.51} & \gain{+3.55}{0.61} & \gain{+3.70}{0.61} \\
\bottomrule
\end{tabular}
\end{table*}

\section{Conclusion and Limitations}
\label{sec:conclusion}

While existing training-free methods effectively enhance the performance of zero-shot CLIP, they 
often underutilize the semantic meaning from text embeddings.
Motivated by this, we introduce DEFTS: a data-efficient, training-free few-shot learning 
method strictly in the text subspace, and show that DEFTS complements
current image-centric methods. We first demonstrate empirically that the text embedding space is highly 
discriminative and data-efficient. DEFTS therefore performs Nearest Class Mean (NCM) estimation directly within this subspace. By expressing image embeddings as linear combinations of text embeddings, DEFTS reduces a high-dimensional estimation problem down to just a few coefficients. These coefficients are computed as a Bayesian posterior mean, utilizing a prior naturally centered on the text embedding coordinates. Extensive experiments demonstrate the effectiveness of DEFTS: by ensembling with
other methods, DEFTS can often improve their accuracy over different datasets.

\paragraph{Limitations} While DEFTS often significantly complements existing baselines, these gains vary by dataset and are sometimes marginal or negative (19 out of 480). Additionally, the relative performance boost provided by DEFTS decreases as the number of shots $k$ increases. This is likely because the baseline methods themselves become substantially stronger with larger support sets. However, even in these higher-shot scenarios, DEFTS demonstrates that it can significantly complement strong baselines such as LDA.

\newpage

\bibliographystyle{plainnat}
\bibliography{references}

\appendix

\makeatletter
\newif\ifapp@aftersection
\app@aftersectionfalse

\pretocmd{\section}{\clearpage\global\app@aftersectiontrue}{}{}

\pretocmd{\subsection}{%
  \FloatBarrier
  \ifapp@aftersection
    \global\app@aftersectionfalse
  \else
    \clearpage
  \fi
}{}{}
\makeatother

\section{More Related Work}
\label{app:related_work}

\rt{
\paragraph{Gradient-based few-shot adaptation of CLIP.} Several methods adapt CLIP with gradient-based optimization. For example, CoOp \citep{zhou2022learning} learns continuous prompt vectors by gradient descent on the labeled few-shot images. \citet{zhao2024test} adapt CLIP at test time with a reward computed from the CLIP score of unlabeled test samples, where they use no labeled support images and update the model at test time. LFA \citep{ouali2023black} learns a linear map from image features to the text embeddings, initialized by a regularized Procrustes solution and refined by gradient-based training.
}

\paragraph{Study of the Shared Embedding Space.} Standard zero-shot CLIP classification relies on cosine similarity, assuming implicitly that image and text embeddings occupy a unified semantic space. However, \citet{liang2022mindthegap} demonstrate a measurable modality gap by showing that image and text embeddings occupy distinctly separated regions on the embedding sphere. This phenomenon mirrors the anisotropy documented in contextual language models \citep{ethayarajh2019contextual, mu2017all}, where representations concentrate their variance into narrow cones. Recent surveys \citep{you2025semantics} and empirical studies \citep{parupudi2026anisotropy, steck2024cosine} reveal that this geometric distortion causes the performances decrease of cosine similarity.

\paragraph{Bayesian Estimation with Small Sample Size.} Estimating a highly parameterized class mean from only a handful of samples is notoriously unstable. Classical statistical theory dictates that the naive per-group sample mean is suboptimal in this regime; instead, shrinking every group's sample mean toward a common point achieves uniformly lower total squared error \citep{james1961estimation}. \citet{efron1975data} operationalized this shrinkage via empirical Bayes, where the strength of the shrinkage is estimated directly from the data's variance rather than heuristically tuned. 

\paragraph{Few-shot Learning.} Few shot learning is highly desirable
in practice where the available training sample size is often limited. 
First, real-world data frequently exhibits 
long-tailed distributions, where a small number of common classes 
dominate the dataset while the vast majority of categories has only a few training samples.
For example, in the diagnosis of rare medical conditions, the detection of 
novel manufacturing defects, or the classification of endangered species,
a class may only be represented by only one or two examples \citep{liu2019large, van2018inaturalist}. 
Second, even when abundant data exists in principle, the cost of expert annotation 
imposes a practical challenge on the volume of data that can be labeled for deployment \citep{esteva2021deep}. 
Third, rapid on-the-fly personalization, such as teaching a robotic assistant
to recognize a user's personal belongings from a single demonstration,
demands models that generalize instantly from extreme few-shot 
interactions \citep{finn2017one}. This capacity for immediate vision-language 
grounding is becoming increasingly central to the deployment of 
real-world robotic manipulation systems \citep{shridhar2022cliport}.

\section{Hyperparameter estimation}
\label{app:hyperparameter}

\subsection{Derivation for the variance of the prior}
\label{app:hyperparameter_partone}

To estimate the hyperparameters without relying on held-out validation data, 
we employ the Empirical Bayes method. 
Let $\bar{b}_c = \frac{1}{k}\sum_{i=1}^k b_{c,i}$ denote the sample mean of 
the projected images for class $c$, and let $\delta_c = \bar{b}_c - e_c$ 
be its deviation from the text prior. For $k \ge 2$, we first isolate the 
observation noise by computing the pooled within-class sample variance across 
all $K$ classes:
\begin{align*}
\hat{\sigma}^2 = \frac{1}{r K(k-1)} \sum_{c=1}^K \sum_{i=1}^k \Vert b_{c,i} - \bar{b}_c \Vert^2.    
\end{align*}
In order to estimate $\tau^2$, we note that the true prior covariance is defined by the 
unobservable true class mean vector $\mu_c \in \mathbb{R}^r$:
\begin{align*}
\text{Cov}(\mu_c - e_c') = \tau^2 I_r.    
\end{align*}
Because we can only measure this distance using the empirical sample mean $\bar{b}_c$, 
we must account for sampling noise. Assume that one sample equals the true mean 
plus a random sampling error vector $\epsilon$. Then we have
\begin{align*}
\bar{b}_c = \mu_c + \frac{1}{k}\sum_{i=1}^k \epsilon_i.    
\end{align*}
Since the true semantic shift and the sampling error are independent, their covariance matrices add linearly:
\begin{align*}
\text{Cov}(\bar{b}_c - e_c') &= \text{Cov}\left(\mu_c + \frac{1}{k}\sum_{i=1}^k \epsilon_i - e_c'\right)\\
&= \text{Cov}(\mu_c - e_c') + \text{Cov}\left(\frac{1}{k}\sum_{i=1}^k \epsilon_i\right) \\
&= \left(\tau^2 + \frac{\sigma^2}{k}\right)I_r.    
\end{align*}

Assuming that $\bar{b}_c$ is an unbiased estimation of $e_c'$, then
taking the expected squared $L_2$ norm equals the trace of the covariance matrix.
Dividing the expected squared $L_2$ norm by the dimensionality $r$, we obtain the per-dimension scalar relationship:
\begin{align*}
\frac{1}{r}\mathbb{E}\left[\Vert\bar{b}_c - e_c'\Vert^2\right] = \tau^2 + \frac{\sigma^2}{k}.    
\end{align*}
Rearranging this equation isolates the true prior variance:
\begin{align*}
\tau^2 = \frac{1}{r}\mathbb{E}\left[\Vert\bar{b}_c - e_c'\Vert^2\right] - \frac{\sigma^2}{k}.    
\end{align*}
Using the observed average squared spread 
\begin{align*}
    \frac{1}{r} \overline{\Vert\delta\Vert^2} = \frac{1}{rK}\sum_{c=1}^K\left(\left\Vert \frac{1}{k}\sum_{i=1}^k b_{c,i} - e_c' \right\Vert^2\right)
\end{align*}
to approximate $\frac{1}{r}\mathbb{E}\left[\Vert\bar{b}_c - e_c'\Vert^2\right]$, 
we derive an estimator of $\tau^2$ by subtracting the expected sampling noise:
\begin{align*}
    \hat{\tau}^2 = \frac{1}{r} \overline{\Vert\delta\Vert^2} - \frac{\hat{\sigma}^2}{k}
\end{align*}
where the overline denotes the average across all $K$ classes.

\paragraph{Remark for the prior assumption.} The derivation above assumes that the text embedding $e_c'$ serves as an unbiased prior for the true image class mean $\mu_c$ (i.e., $\mathbb{E}[\mu_c - e_c'] = 0$). While recent literature has identified a systematic ``modality gap'' between image and text embeddings in contrastive models \citep{liang2022mindthegap}, explicitly modeling this shift requires estimating a high-dimensional bias vector. In few-shot regimes (e.g., $k \in \{1, 2, 4\}$), such estimation could be prone to overfitting. We therefore deliberately enforce this zero-bias assumption as a regularization constraint.

\subsection{Derivation for the shrunk covariance estimation}
\label{app:hyperparameter_parttwo}

In this section, we show the derivation of the 
Bayesian shrinkage estimator for the covariance matrix.

Let $\hat{\Sigma}$ denote the pooled empirical sample covariance computed from $K$ classes and $k$ support shots. The total degrees of freedom for this empirical estimate is 
\begin{align*}
    n = K(k-1).
\end{align*}
We define the empirical scatter matrix $S$ as the 
unnormalized sum of squared errors: $S = n\hat{\Sigma}$.
Under the assumption that the image embeddings are drawn from a Multivariate Normal distribution $\mathcal{N}(\mu_c, \Sigma)$, this scatter matrix follows a Wishart distribution, $S \sim \mathcal{W}_r(\Sigma, n)$ \citep{murphy2012machine}. 
The data likelihood for the covariance matrix is therefore proportional to:
\begin{align*}
    p(S \mid \Sigma) \propto \vert{}\Sigma\vert{}^{-n/2} \exp\left(-\frac{1}{2}\operatorname{tr}(\Sigma^{-1}S)\right).
\end{align*}

To regularize $\hat{\Sigma}$ in the few-shot cases, 
we place an Inverse-Wishart prior on the true covariance matrix $\Sigma$:
\begin{align*}
    \Sigma \sim \mathcal{IW}_r(\Psi, \nu_0),
\end{align*}
where $\Psi$ is the $r \times r$ scale matrix and $\nu_0$ represents the prior degrees of freedom. Following an Empirical Bayes approach to covariance shrinkage \citep{ledoit2004well}, we set the prior scale matrix $\Psi$ to be an isotropic target matrix that preserves the total variance of the observed data
\begin{align*}
    \Psi = \nu_0 \hat{\sigma}^2 I_r = \nu_0 \frac{\operatorname{tr}(\hat{\Sigma})}{r} I_r.
\end{align*}

As shown by \citet{murphy2012machine}, the posterior distribution $p(\Sigma \mid S) \propto p(S \mid \Sigma)p(\Sigma)$ is also an Inverse-Wishart distribution such that
\begin{align*}
    p(\Sigma \mid S) \propto \vert{}\Sigma\vert{}^{- \frac{n + (\nu_0 + r + 1)}{2}} \exp\left(-\frac{1}{2}\operatorname{tr}(\Sigma^{-1}(S + \Psi))\right).
\end{align*}
Following the posterior distribution, the maximum a posteriori (MAP) estimation of $\Sigma$ is:
\begin{align*}
    \Sigma = \frac{S + \Psi}{n + \nu_0 + r + 1}
\end{align*}
Substituting our definitions of $S$ and $\Psi$ yields:
\begin{align*}
S + \Psi = n\hat{\Sigma} + \nu_0 \frac{\operatorname{tr}(\hat{\Sigma})}{r} I_r.    
\end{align*}
Setting $\nu_0 = r$, we further have
\begin{align*}
S + \Psi = n\hat{\Sigma} + \operatorname{tr}(\hat{\Sigma})I_r.    
\end{align*}

Dividing this combined scatter matrix by the total effective degrees of freedom
yields the standard Bayes-shrunk covariance. However, in few shot learning regimes, especially
when the number of shots $k$ is extremely small, e.g. $k = 2$,
the covariance $\hat{\Sigma}$ could be highly unstable as shown in Appendix~\ref{app:covariance}. 
We therefore introduce a covariance inflation parameter $\beta \in (0, 1]$, explicitly inflating the estimated variance by a factor of $\frac{1}{\beta}$:
\begin{align*}
    \Sigma = \frac{1}{\beta} \frac{n\hat{\Sigma} + \operatorname{tr}(\hat{\Sigma})I_r}{n + 2r + 1}.
\end{align*}
Substituting $n = K(k-1)$ and taking the matrix inverse yields our final regularized precision matrix:
\begin{align*}
    \Sigma^{-1} = \beta (K(k-1) + 2r + 1) \left[ K(k-1)\hat{\Sigma} + \operatorname{tr}(\hat{\Sigma})I_r \right]^{-1}.
\end{align*}

\subsection{\rt{The one-shot estimator and the role of the Bayesian update}}
\label{app:oneshot}

\rt{At $k=1$ the within-class variance cannot be estimated, and DEFTS uses $\Sigma^{-1}=\beta\hat\sigma^{-2}I_r$, $\hat\sigma^2=f\,s$ and $\hat\tau^2=(1-f)\,s$, where $s=\frac1r\lVert\delta\rVert^2$ is the observed spread and $f$ is the noise fraction. We now derive the estimator of DEFTS at $k=1$. }

\rt{According to Equation~\ref{eqn:posterior}, we have
\begin{align*}
    \hat{\theta}_c = (\Omega_0 + k\Sigma^{-1})^{-1} (\Omega_0 e_c' + k\Sigma^{-1} \bar{b}_c),
\end{align*}
where $\Omega_0 = \hat{\tau}^{-2} I_r$. Then, let $k = 1$, we have
\begin{align*}
    \hat{\theta}_c &= (\hat{\tau}^{-2} I_r + \beta\hat\sigma^{-2}I_r)^{-1}(\hat{\tau}^{-2} I_r e_c' + \beta\hat\sigma^{-2}I_r \bar{b}_c) \\
    & = \frac{\hat\tau^{-2}\,e'_c+\beta\hat\sigma^{-2}\,\bar b_c}{\hat\tau^{-2}+\beta\hat\sigma^{-2}} =(1-w_1)\,e'_c+w_1\,\bar b_c,
\end{align*}
where $w_1={\beta\hat\sigma^{-2}}/{(\hat\tau^{-2}+\beta\hat\sigma^{-2})}$. Substituting $\hat\sigma^{-2}=1/(fs)$ and $\hat\tau^{-2}=1/((1-f)s)$,
\begin{equation*}
w_1=\frac{\beta/(fs)}{1/((1-f)s)+\beta/(fs)}=\frac{\beta/f}{1/(1-f)+\beta/f}=\frac{\beta(1-f)}{f+\beta(1-f)},
\end{equation*}
where $s$ cancels. For $\beta=0.1$ and $f=0.1$, $w_1=0.090/0.190\approx0.474$.}

\newpage
\section{DEFTS algorithm}
\label{app:algorithm}

In this section, we provide the algorithm for
DEFTS in Algorithm~\ref{alg:defts}, which summarizes 
our discussions about DEFTS in Section~\ref{sec:defts}

\begin{algorithm}[ht]
\caption{DEFTS Prototype Generation and Inference}
\label{alg:defts}
\begin{algorithmic}[1] 
\Require $K$ text embeddings $\{e_c\}_{c=1}^K$, $k$ support image embeddings per class $\{x_{c,i}\}_{i=1}^k$
\Ensure Estimated image prototypes $\{\hat{\mu}_c\}_{c=1}^K$ for classification

\State Compute the centralized scatter matrix $S$ from the text embeddings $\{e_c\}_{c=1}^K$.
\State Eigendecompose $S$ to obtain the PCA basis $Q \in \mathbb{R}^{d \times r}$, choosing $r$ to retain 95\% of the variance.

\For{each class $c \in \{1, \dots, K\}$}
    \State Project text embedding: $e_c' \gets Q^\top e_c$
    \State Project support images: $b_{c,i} \gets Q^\top x_{c,i} \quad \forall i \in \{1, \dots, k\}$
\EndFor

\State Estimate the prior variance $\hat{\tau}^2$ and the pooled likelihood precision $\Sigma^{-1}$ directly from the projected $k$-shot support set $\{b_{c,i}\}$ via Empirical Bayes as shown in \eqref{eqn:sigma} and \eqref{eqn:tau}.

\For{each class $c \in \{1, \dots, K\}$}
    \State Compute the posterior mean $\hat{\theta}_c$ in the subspace following Equation~\eqref{eqn:posterior}.
    \State Reconstruct the high-dimensional prototype: $\hat{\mu}_c \gets Q \hat{\theta}_c$.
\EndFor

\State During inference, for a query image $I$, compute the logit for class $c$ as:
\begin{align*}
    s_{\mathrm{DEFTS}}(F_{v}(I))_c =  \frac{F_{v}(I)^\top}{\lVert F_{v}(I)\rVert}\frac{\hat{\mu}_c}{\lVert \hat{\mu}_c \rVert} .
\end{align*}

\end{algorithmic}
\end{algorithm}

\newpage

\section{Structural Comparison Between DEFTS and TAMP}
\label{sec:defts_vs_tamp}

Although both DEFTS and TAMP~\citep{goswami2026cross} exploit the text embedding space, they employ fundamentally different estimation mechanisms. TAMP uses the text space primarily as a geometric projection space and combines the projected visual prototype with the text prototype via a scalar interpolation coefficient. In contrast, DEFTS treats the text-derived subspace as a statistical estimation space, performing uncertainty-aware, direction-dependent Bayesian shrinkage. We formalize this distinction below.

Let $\bar{x}_c$ denote the empirical mean of the $k$-shot support image embeddings for class $c$ in the ambient space, and let $e_c$ denote the corresponding text embedding. TAMP constructs an orthonormal basis $Q' \in \mathbb{R}^{d\times r}$ from the uncentered text embeddings and defines the corresponding projection matrix as
\begin{align*}
    P = Q'Q'^\top.
\end{align*}
Since $e_c$ lies in the span of $Q'$, we have $Q'Q'^\top e_c=e_c$. TAMP then forms its class prototype by interpolating between the projected image prototype and the text prototype:
\begin{align}
W_{\mathrm{TAMP}}
&= \lambda Q'Q'^\top \bar{x}_c + (1-\lambda)e_c \nonumber\\
&= Q' \left[
\lambda I_r(Q'^\top\bar{x}_c) +
(1-\lambda)I_r(Q'^\top e_c)
\right].
\label{eq:tamp_structural}
\end{align}
Thus, within the text subspace, TAMP applies the same scalar weight to every direction. The relative contribution of the visual estimate and the text prototype is therefore controlled by a single global coefficient $\lambda$.

DEFTS, instead, constructs a centered PCA basis $Q\in\mathbb{R}^{d\times r}$ from the text embeddings and performs Bayesian estimation in the resulting subspace. Define
\begin{align*}
    \bar{b}_c = Q^\top\bar{x}_c, \qquad e'_c = Q^\top e_c.
\end{align*}
Following Eq.~\eqref{eqn:posterior}, the posterior mean in the PCA coordinates is
\begin{align*}
    \hat{\theta}_c = 
\left(
\Omega_0+ k\Sigma^{-1}
\right)^{-1}
\left(
 k\Sigma^{-1}\bar{b}_c
+
\Omega_0 e'_c
\right),
\end{align*}
and the corresponding prototype in the original embedding space is
\begin{align}
W_{\mathrm{DEFTS}}
&=
Q\hat{\theta}_c \nonumber\\
&=
Q
\left[ \left( \Omega_0+ k\Sigma^{-1}\right)^{-1}  k\Sigma^{-1} (Q^\top\bar{x}_c)
+ \left( \Omega_0+ k\Sigma^{-1} \right)^{-1} \Omega_0 (Q^\top e_c) \right].
\label{eq:defts_structural}
\end{align}
For notational convenience, let
\begin{align*}
    A= \left( \Omega_0+ k\Sigma^{-1} \right)^{-1}, \qquad B= k\Sigma^{-1}, \qquad C=\Omega_0.
\end{align*}
Then
\begin{equation}
W_{\mathrm{DEFTS}} = 
Q
\left[
AB(Q^\top\bar{x}_c)
+
AC(Q^\top e_c)
\right].
\label{eq:defts_structural_compact}
\end{equation}

These expressions reveal three important structural differences.

\paragraph{Scalar interpolation versus uncertainty-aware anisotropic fusion.}
TAMP uses a scalar coefficient $\lambda$ and the identity matrix $I_r$, effectively treating all directions within the text subspace identically. DEFTS, instead, weights the visual observation and text prior through the dense matrices $AB$ and $AC$, which depend on the estimated covariance $\Sigma$ and prior precision $\Omega_0$. Consequently, the relative contribution of the image evidence and the text prior varies across directions based on their estimated uncertainty. Directions with lower visual variance receive relatively greater weight from the empirical image evidence, whereas high-variance directions are shrunk more strongly toward the text prior. DEFTS therefore performs uncertainty-aware, direction-dependent prototype estimation rather than a globally weighted interpolation.

This distinction also provides a useful interpretation of scalar interpolation. If the uncertainty structure were isotropic, i.e.,
\[
\Sigma=\sigma^2 I_r
\quad\text{and}\quad
\Omega_0=\omega^2 I_r,
\]
the corresponding Bayesian fusion would reduce exactly to a scalar weighting of the visual estimate and the prior. Thus, scalar interpolation can be viewed as the isotropic special case of the precision-weighted fusion employed by DEFTS. DEFTS extends this formulation by allowing the estimated uncertainty to dynamically differ across directions.

\paragraph{Scalar hyperparameter selection versus Empirical Bayes estimation.}
TAMP determines the relative contribution of the projected image prototype and text prototype through the scalar coefficient $\lambda$, which must be externally tuned via validation. DEFTS estimates the visual uncertainty structure directly from the available training data using the Empirical Bayes procedure. The resulting precision matrix dictates how strongly the image evidence and text prior are combined along each axis, thus avoiding the need for searching hyperparameter for each dataset.

\paragraph{Uncentered text span versus centered PCA subspace.}
The two methods also differ in their chosen representation space. TAMP constructs $Q'$ directly from the uncentered text embeddings, operating within their uncentered linear span. DEFTS first centers the text embeddings before constructing a PCA basis that retains a fixed fraction of their variance. Empirically, our ablation in Appendix~\ref{app:pcacenter} suggests that the centered construction yields more effective classification, which is why it forms the foundation of DEFTS.

Overall, while TAMP and DEFTS share the intuition that text embeddings provide a highly structural low-dimensional space for few-shot visual classification, they exploit it differently. TAMP applies a uniform scalar interpolation after projecting the visual prototype into the text space, whereas DEFTS formulates prototype estimation as uncertainty-aware Bayesian shrinkage within a centered PCA subspace. This capability to weight visual evidence and text priors differently across dimensions is critical in few-shot cases, where visual prototype estimates remain highly uncertain. We summarize the differences between TAMP and DEFTS in Table~\ref{tab:method_comparison}.

\begin{table}[htbp]
\centering
\caption{Key structural and methodological differences between the TAMP baseline and the proposed DEFTS framework.}
\label{tab:method_comparison}
\renewcommand{\arraystretch}{1.2} 
\begin{tabular}{@{}lll@{}}
\toprule
\textbf{Component} & \textbf{TAMP} & \textbf{DEFTS} \\
\midrule
Image representation & Projected image mean & Projected image mean \\
Text representation & Text prototype & Text prototype as prior mean \\
Subspace & Uncentered text span & Centered varaince reduced PCA subspace \\
Fusion & Scalar interpolation & Precision-weighted fusion \\
Image weighting & Global $\lambda$ & Direction-dependent \\
Uncertainty & Not modeled explicitly & Estimated covariance \\
Estimation & Validation-selected $\lambda$ & Empirical Bayes \\
Output & Interpolated prototype & Posterior-mean prototype \\
\bottomrule
\end{tabular}
\end{table}

\newpage
\section{More Experiments details}
\label{app:details}

In this section, we provide more details for the experiments in Section~\ref{sec:textsubspace}
and~\ref{sec:experiments}.

\subsection{Datasets Details}
\label{app:dataset}
In Table~\ref{tab:dataset_statistics}, we provide more details about the 12 datasets considered in this paper, including the total number of images, the number of classes, and the minimum and maximum images per class.

\begin{table*}[h]
\centering
\caption{Summary of the 12 evaluation datasets, detailing the total number of images, the number of classes, and the minimum and maximum images per class.}
\label{tab:dataset_statistics}
\begin{tabular}{l rr rr}
\toprule
\textbf{Dataset} & \textbf{Total Images} & \textbf{\# Classes} & \textbf{Min/Class} & \textbf{Max/Class} \\
\midrule
CIFAR10       & 10,000 & 10    & 1,000 & 1,000 \\
CIFAR100      & 10,000 & 100   & 100   & 100   \\
EuroSAT       & 27,000 & 10    & 2,000 & 3,000 \\
Food101       & 25,250 & 101   & 250   & 250   \\
OxfordIIITPet & 3,669  & 37    & 88    & 100   \\
FGVCAircraft  & 3,333  & 100   & 33    & 34    \\
Flowers102    & 8,189  & 102   & 40    & 258   \\
DTD           & 1,880  & 47    & 40    & 40    \\
ImageNet      & 50,000 & 1,000 & 50    & 50    \\
Caltech101    & 8,242  & 100   & 31    & 800   \\
StanfordCars  & 8,041  & 196   & 24    & 68    \\
SUN397        & 19,850 & 397   & 50    & 50    \\
\bottomrule
\end{tabular}
\end{table*}

\subsection{Prompt Templates}
\label{app:prompt}

In Section~\ref{sec:experiments}, we evaluate the performance of DEFTS using the standard prompt templates introduced by CoOp \citep{zhou2022learning}. This section details the exact text prompt applied to each corresponding dataset.

\begin{table}[ht]
\centering
\caption{Text prompt templates utilized for each of the 12 evaluation datasets. The placeholder \{c\} denotes the specific class name.}
\label{tab:prompt_templates}
\begin{tabular}{ll}
\toprule
\textbf{Dataset} & \textbf{Prompt Template} \\
\midrule
CIFAR10 & a photo of a \{c\}. \\
CIFAR100 & a photo of a \{c\}. \\
EuroSAT & a centered satellite photo of \{c\}. \\
Food101 & a photo of \{c\}, a type of food. \\
OxfordIIITPet & a photo of a \{c\}, a type of pet. \\
FGVCAircraft & a photo of a \{c\}, a type of aircraft. \\
Flowers102 & a photo of a \{c\}, a type of flower. \\
DTD & \{c\} texture. \\
ImageNet & a photo of a \{c\}. \\
Caltech101 & a photo of a \{c\}. \\
StanfordCars & a photo of a \{c\}. \\
SUN397 & a photo of a \{c\}. \\
\bottomrule
\end{tabular}
\end{table}

\subsection{Subspace ranks}
\label{app:ranks}

\rt{DEFTS proposes to retain 95\% of the variance of the centered PCA for each dataset. Table~\ref{tab:ranks} reports the number of classes $K$ and the resulting subspace dimension $r$ used by DEFTS across all datasets and the three different vision backbones.}

\begin{table}[ht]
\centering
\color{blue}
\caption{\rt{Rank of the centered basis keeping 95\% of the text-embedding variance, for three different model backbones: ViT-L/14, ViT-B/32 and ViT-H/14.}}
\label{tab:ranks}
\small
\begin{tabular}{@{}l r rrr@{}}
\toprule
Dataset & $K$ & $r$ (ViT-L/14) & $r$ (ViT-B/32) & $r$ (ViT-H/14) \\
\midrule
CIFAR10 & 10 & 9 & 8 & 9 \\
CIFAR100 & 100 & 75 & 73 & 74 \\
Caltech101 & 100 & 75 & 73 & 74 \\
DTD & 47 & 37 & 35 & 38 \\
EuroSAT & 10 & 8 & 8 & 8 \\
FGVCAircraft & 100 & 46 & 42 & 46 \\
Flowers102 & 102 & 72 & 68 & 68 \\
Food101 & 101 & 73 & 71 & 73 \\
Imagenet & 1000 & 289 & 241 & 289 \\
OxfordIIITPet & 37 & 27 & 25 & 26 \\
SUN397 & 397 & 182 & 166 & 184 \\
StanfordCars & 196 & 79 & 74 & 82 \\
\bottomrule
\end{tabular}
\end{table}

\subsection{Experiments Details for Observations}
\label{app:experimentsobs}

In Section~\ref{sec:textsubspace}, we present 3 observations 
for investigating the text embedding subspace. In this section,
we provide details for the experiments supporting our observations.

\paragraph{Experiments for Observation 1.} For Observation 1, 
we train an NCM classifier for each dataset. We first randomly sample 80\% of the images 
from each class as training data. Then, we project all image features into the text subspace, as detailed in Section~\ref{sec:bayes-pca} but without centering first. The mean of the image embeddings within this subspace are estimated by taking average and used for performing nearest class mean classification. The classification accuracy is evaluated on the remaining 20\% of the images, essentially serving as the validation set. This procedure is repeated over 10 independent runs with varying random seeds, and the final reported accuracy is the average across all runs.

\paragraph{Experiments for Observation 2.} The experimental setup for Observation 2 
is similar to that for Observation 1. However, in this experiment, 
we randomly sample $k$ images from each class to serve as the training data, 
while using all remaining images as the validation set.

\paragraph{Experiments for Observation 3.} Similarly, the experiment for Observation 3 
builds upon the one for Observation 1. The difference is that we now first reduce the 
dimensionality of the subspace, as described in Section~\ref{sec:bayes-pca}.

\newpage
\section{Additional Experiments Results}
\label{app:additional}

This section includes additional experiments results.

\subsection{Text Subspace with Different Template}
\label{app:template}

This additional experiment provides evidence that the observations 
discussed in Section~\ref{sec:textsubspace} are robust to the choice 
of prompt template. To this end, we perform the same experiments
for the discriminative power of text embedding space in Figure~\ref{fig:subspace_comparison}
with different prompt templates.

\begin{figure}[h]
\centering
\includegraphics[width=1\linewidth]{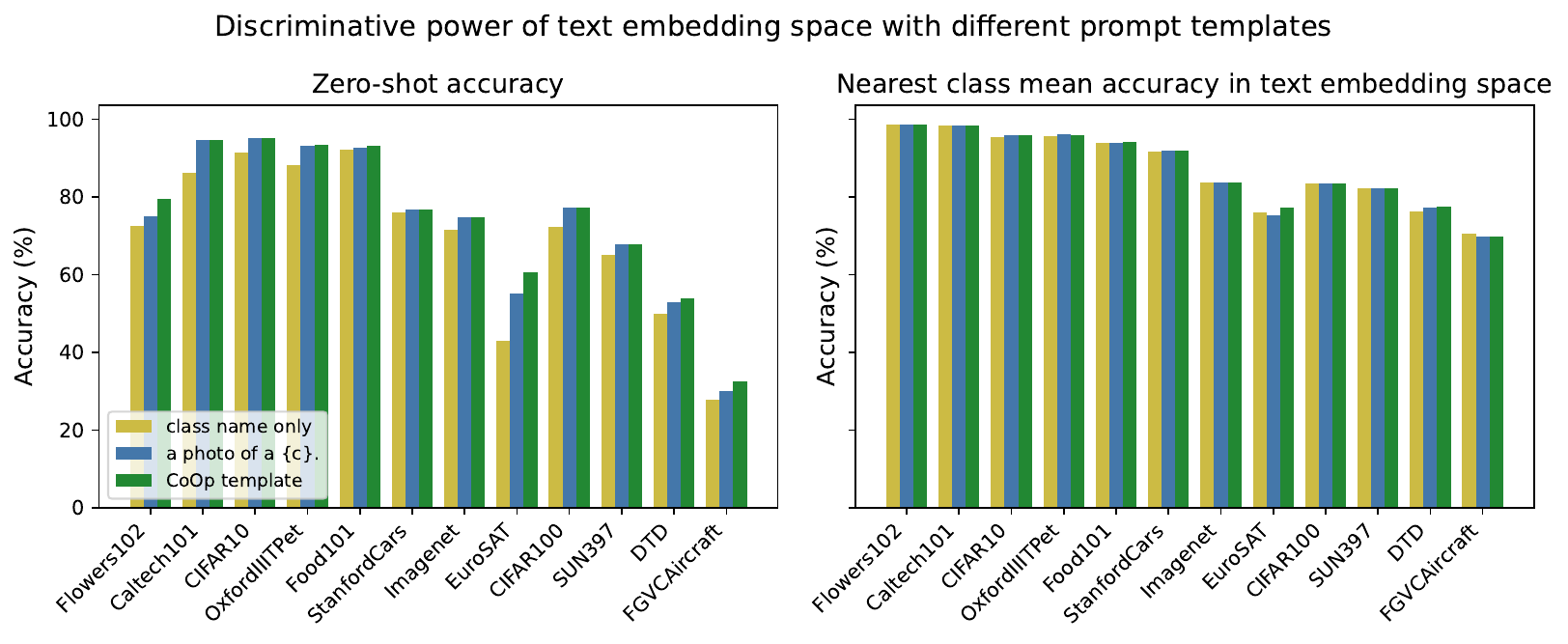}
\caption{CLIP's zero-shot accuracy and Nearest Class Mean (NCM) accuracy across three different prompt templates. To compute the NCM, all image embeddings are first projected into the text-embedding subspace to estimate the mean embedding for each class. A test image is then classified by assigning it to the class with the closest estimated mean.}
\label{fig:templates}
\end{figure}

Figure~\ref{fig:templates} evaluates three different prompt templates: 
(1) the class name alone; (2) a standard fixed template ``a photo of a \{class\}''; 
and (3) the hand-crafted templates utilized in CoOp \citep{zhou2022learning}, as detailed in Appendix~\ref{app:prompt}.
As expected, improving the template quality steadily increases CLIP's baseline zero-shot accuracy. However, the results provide evidence suggesting that
the template quality has a more negligible impact on the discriminative power of the text embedding space itself.

\subsection{Subspace advantage, \texorpdfstring{$K/d$}{K/d} and the number of classes}
\label{app:kd}

{\color{blue}
Observation 2 shows that the prototype estimation in the text subspace is usually 
more data-efficient than in the full space. In this section, we investigate if this 
advantage correlates with the $K/d$ ratio. Table~\ref{tab:kd_advantage} considers the relative 
gains across different datasets and number of shots $k$ and calculates the corresponding 
Spearman's rank correlation. The results show no statistically significant relationship
between the advantage and the $K/d$ ratio at the 5\% level.}

\begin{table}[!ht]
\centering
\color{blue}
\caption{{\color{blue}{Advantage of the text span over the full space (text span NCM minus full-space NCM) for each dataset and number of shots $k$, with datasets ordered by $K/d$. The last two rows give Spearman's $\rho$ between the advantage and $K/d$ over the 12 datasets and its $p$-value. The results are obtained using model backbone ViT-L/14 and are averaged over 10 random seeds.}}}
\label{tab:kd_advantage}
{\small
\renewcommand{\arraystretch}{1.1}
\begin{tabular}{@{}l r r ccccc@{}}
\toprule
\textbf{Dataset} & $K$ & $K/d$ & $k=1$ & $k=2$ & $k=4$ & $k=8$ & $k=16$ \\
\midrule
CIFAR10 & 10 & 0.013 & $+$30.4 & $+$14.9 & $+$5.2 & $+$2.6 & $+$0.8 \\
EuroSAT & 10 & 0.013 & $+$5.0 & $-$0.2 & $-$4.4 & $-$6.6 & $-$6.5 \\
Pets & 37 & 0.048 & $+$36.3 & $+$22.3 & $+$12.9 & $+$6.6 & $+$4.0 \\
DTD & 47 & 0.061 & $+$12.1 & $+$9.4 & $+$7.6 & $+$5.9 & $+$4.9 \\
CIFAR100 & 100 & 0.130 & $+$23.2 & $+$20.5 & $+$13.9 & $+$8.9 & $+$6.3 \\
Caltech101 & 100 & 0.130 & $+$11.6 & $+$4.1 & $+$1.8 & $+$1.1 & $+$1.0 \\
Aircraft & 100 & 0.130 & $+$8.0 & $+$6.8 & $+$6.0 & $+$4.8 & $+$3.6 \\
Food101 & 101 & 0.132 & $+$16.8 & $+$10.9 & $+$5.4 & $+$3.1 & $+$2.2 \\
Flowers102 & 102 & 0.133 & $+$5.6 & $+$1.7 & $+$0.4 & $+$0.1 & $+$0.1 \\
Cars & 196 & 0.255 & $+$17.7 & $+$14.2 & $+$10.4 & $+$7.7 & $+$6.3 \\
SUN397 & 397 & 0.517 & $+$6.3 & $+$4.6 & $+$3.0 & $+$2.0 & $+$1.5 \\
ImageNet & 1000 & 1.302 & $+$0.0 & $+$0.0 & $+$0.0 & $+$0.0 & $+$0.0 \\
\midrule
Spearman $\rho$ with $K/d$ & & & $-$0.43 & $-$0.29 & $-$0.15 & $-$0.10 & $+$0.00 \\
$p$-value & & & 0.17 & 0.37 & 0.64 & 0.75 & 0.99 \\
\bottomrule
\end{tabular}}
\end{table}

\subsection{Centered PCA Projection}
\label{app:pcacenter}

In Section~\ref{sec:defts}, DEFTS proposes to project image embeddings into a centered PCA subspace,
rather than the original, uncentered subspace. To justify this design choice, we perform an ablation study comparing centered versus uncentered PCA projections while retaining either 95\% or 99.9\% of the total variance. Table~\ref{tab:pca_centered} suggests that for 95\% variance kept, 
the centered PCA wins uncentered one by larger margin, around 2-3\%,
when 99.9\% variance are kept, centered one is often slightly better
than uncentered one.

\begin{table}[ht]
\centering
\caption{DEFTS's average accuracy (\%) across all 12 datasets for centered and uncentered PCA variants across 95\% and 99.9\% explained variance thresholds for varying support set sizes ($k$). The best performance for each $k$ is highlighted in \textbf{\textcolor{red}{bold red}}. Results are averaged over 10 random seeds and reported as percentages.}
\label{tab:pca_centered}
\begin{tabular}{l cccc}
\toprule
\multirow{2}{*}{$\boldsymbol{k}$} & \multicolumn{2}{c}{\textbf{95\% Variance}} & \multicolumn{2}{c}{\textbf{99.9\% Variance}} \\
\cmidrule(lr){2-3} \cmidrule(lr){4-5}
& \textbf{Centered} & \textbf{Uncentered} & \textbf{Centered} & \textbf{Uncentered} \\
\midrule
1  & \textbf{\textcolor{red}{76.97}} & 73.54 & 76.63 & 76.73 \\
2  & 79.60 & 76.65 & \textbf{\textcolor{red}{79.72}} & 79.56 \\
4  & \textbf{\textcolor{red}{81.77}} & 79.25 & 81.68 & 81.64 \\
8  & \textbf{\textcolor{red}{83.59}} & 81.37 & 83.48 & 83.27 \\
16 & 84.68 & 82.51 & \textbf{\textcolor{red}{84.70}} & 84.41 \\
\bottomrule
\end{tabular}
\end{table}

\subsection{Empirical contribution of total average deviation}
\label{app:contribution}

In this experiment, we investigate the contribution
of the sample variance $\hat{\sigma}^2$ to the averaged deviation 
$\frac{1}{r} \overline{\Vert\delta\Vert^2}$.
Table~\ref{tab:k_averages} shows the results
of the ratio of the sample variance $\hat{\sigma}^2$ to the averaged empirical deviation $\frac{1}{r} \overline{\Vert\delta\Vert^2}$ across 12 datasets for different number of shots $k$. It suggests that this ratio is empirically ranges from $2\%$ to $16\%$, varying across different datasets, but more stable with different number of shots $k$.

\begin{table}[ht]
\centering
\caption{Ratio of the sample variance $\hat{\sigma}^2$ to the averaged empirical deviation $\frac{1}{r} \overline{\Vert\delta\Vert^2}$ across 12 datasets for different number of shots $k$. Results are averaged over 10 random seeds and reported as percentages.}
\label{tab:k_averages}
\begin{tabular}{l ccccc}
\toprule
\multirow{2}{*}{\textbf{Dataset}} & \multicolumn{5}{c}{\textbf{Number of shots ($k$)}} \\
\cmidrule(lr){2-6}
& $\boldsymbol{k=2}$ & $\boldsymbol{k=4}$ & $\boldsymbol{k=8}$ & $\boldsymbol{k=16}$ & \textbf{Average} \\
\midrule
CIFAR10 & 5.09 & 4.94 & 3.81 & 5.21 & 4.76 \\
CIFAR100 & 5.97 & 6.00 & 6.12 & 6.14 & 6.06 \\
Caltech101 & 6.81 & 6.61 & 6.43 & 6.52 & 6.59 \\
DTD & 11.16 & 11.47 & 11.04 & 10.95 & 11.16 \\
EuroSAT & 2.03 & 2.82 & 3.23 & 3.22 & 2.83 \\
FGVCAircraft & 7.46 & 7.83 & 7.69 & 7.72 & 7.67 \\
Flowers102 & 5.31 & 5.70 & 5.38 & 5.37 & 5.44 \\
Food101 & 11.42 & 11.69 & 12.04 & 12.15 & 11.82 \\
ImageNet & 16.27 & 16.00 & 16.07 & 16.11 & 16.11 \\
OxfordPet & 3.24 & 3.20 & 3.42 & 3.24 & 3.27 \\
SUN397 & 14.52 & 14.87 & 14.61 & 14.71 & 14.68 \\
StanfordCars & 8.91 & 9.20 & 8.99 & 8.98 & 9.02 \\
\midrule
\textbf{Average} & \textbf{8.18} & \textbf{8.36} & \textbf{8.24} & \textbf{8.36} & \textbf{8.28} \\
\bottomrule
\end{tabular}
\end{table}

\rt{Thus, when $k = 1$, where $\sigma^2$ and $\tau^2$ are unidentifiable, we set
set $\hat{\sigma}^2 = f\frac{1}{r} \overline{\Vert\delta\Vert^2}$, $\hat{\tau}^2 = (1-f)\frac{1}{r} \overline{\Vert\delta\Vert^2}$ with $f = 1$.}

\begin{figure}[ht]
\centering
\includegraphics[width=0.62\linewidth]{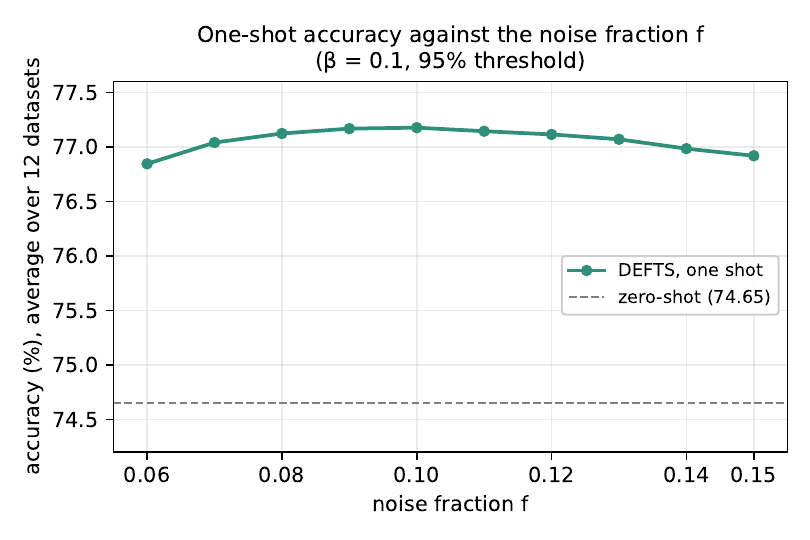}
\caption{\rt{Average one-shot accuracy (\%) over 12 datasets against the noise fraction $f$ for $f=0.06,0.07,\dots,0.15$ with $\beta=0.1$ and 95\% PCA variance threshold. The results are obtained using model backbone ViT-L/14 and are averaged over 10 random seeds. The dashed line shows the accuracy of zero-shot CLIP.}}
\label{fig:fsweep}
\end{figure}

\rt{In Figure~\ref{fig:fsweep}, we show the one-shot accuracy for noise fractions $f$ from $0.06$ to $0.15$ in steps of $0.01$. The results suggest that the accuracy is robust against different choices of the fraction in the range $[0.06, 0.15]$.}

\subsection{Covariance Estimation for Different Number of Shots}
\label{app:covariance}

\begin{figure}[ht]
\centering
\includegraphics[width=1\linewidth]{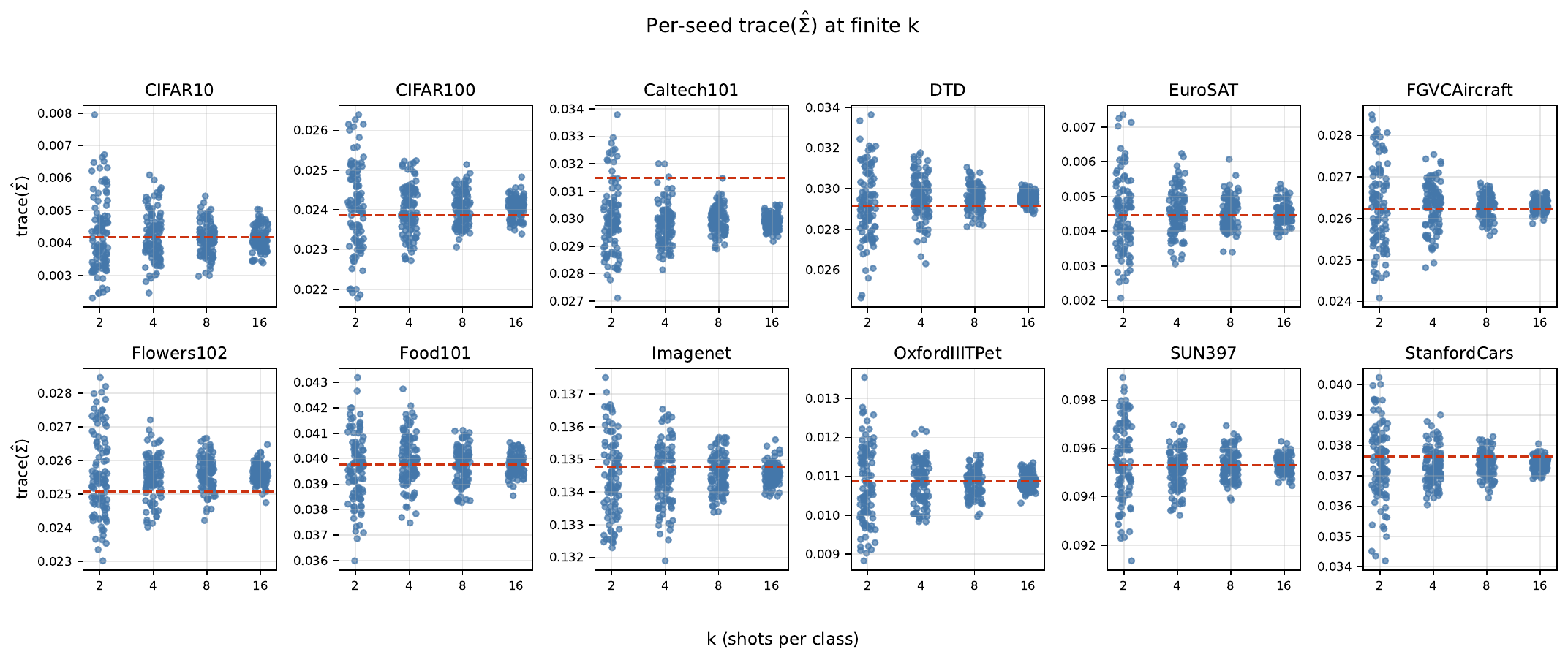}
\caption{Trace of the empirical sample covariance matrix $\hat{\Sigma}$ across 12 datasets for different number of shots $k$. Blue dots represent individual estimates across 100 random seeds, while the red dashed line represents the true population trace estimated using all samples.}
\label{fig:beta1}
\end{figure}

In Section~\ref{sec:defts}, we introduce a covariance inflation
parameter $\beta$ to account for the fact that the covariance
estimation is often problematic especially when the number of shots
$k$ is small. In this section, we presents experiments results
in Figure~\ref{fig:beta1} demonstrating that when the number of shots
is small, the estimation of the covariance could be highly unstable.

\subsection{Impact of the covariance inflation parameter $\beta$ for DEFTS}
\label{app:beta}

\begin{figure}[ht]
\centering
\includegraphics[width=0.5\linewidth]{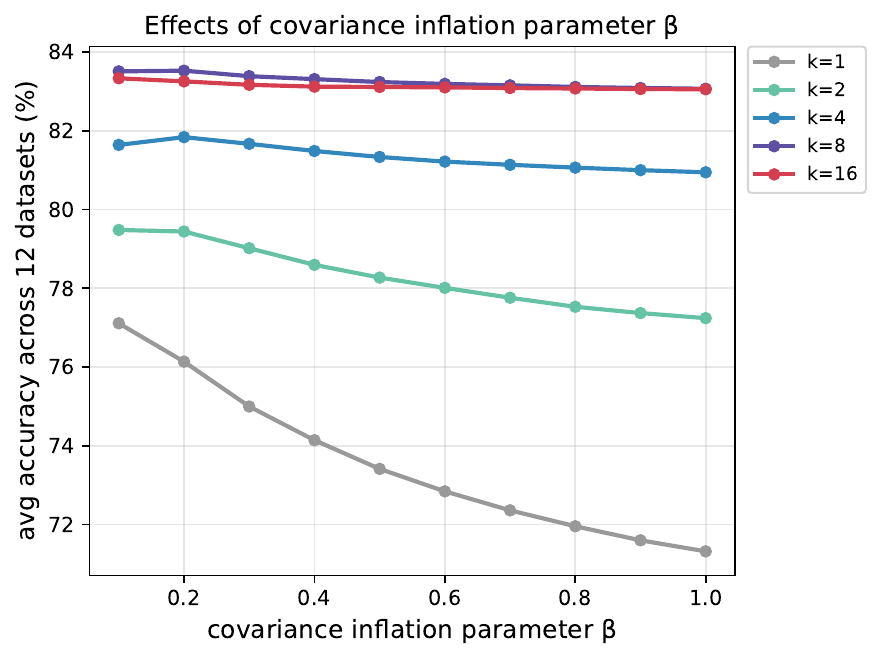}
\caption{DEFTS' average accuracy with different covariance inflation parameter $\beta$ across 12 datasets.}
\label{fig:beta}
\end{figure}

In this additional experiments, we investigate the impact of the  covariance inflation
parameter $\beta$ on the accuracy of DEFTS. In Figure~\ref{fig:beta}, we present DEFTS' average accuracy with different  $\beta$ across 12 datasets. The results suggest that when the number of shots $k$ is small, it is preferable to choose small $\beta$ for DEFTS. However, when the number of shots is larger, the choice of $\beta$ is less critical.

\subsection{PCA of the text embedding space}
\label{app:pca-variance}

In \textbf{Observation 3}, we provide evidence showing that
the discriminative power of the text embedding space
will decrease if we reduce the dimension of the space using PCA.
Note that the discriminative power in \textbf{Observation 3}
is evaluated using NCM with image embeddings estimated using all available
training samples.
In this additional experiment, we use only a few shots of training samples and
evaluate the nearest class mean (NCM) accuracy 
after removing different amounts of variance from the text embedding space.
The results in Table~\ref{tab:pca-variance-average} suggests that for a given 
number of shots $k$, reducing the variance of the text embedding space by 
5\% yields the highest NCM accuracy within the text subspace.

\begin{table}[!htbp]
\centering
\caption{Nearest class mean (NCM) accuracy after reducing different amount 
of variance of the text embedding space. Bold marks the best representation at each $k$.}
\label{tab:pca-variance-average}
\begin{tabular}{lccccc}
\toprule
 & $k=1$ & $k=2$ & $k=4$ & $k=8$ & $k=16$ \\
\midrule
raw (no projection) & 50.86 & 64.99 & 74.14 & 79.17 & 81.82 \\
text span (full) & 65.26 & 74.13 & 79.36 & 82.21 & 83.82 \\
PCA 95\% variance & \textbf{67.87} & \textbf{75.95} & \textbf{80.40} & \textbf{82.85} & \textbf{84.23} \\
PCA 90\% variance & 67.71 & 75.74 & 80.17 & 82.60 & 83.93 \\
PCA 85\% variance & 67.03 & 75.20 & 79.62 & 82.09 & 83.48 \\
PCA 80\% variance & 66.23 & 74.38 & 78.91 & 81.45 & 82.85 \\
PCA 75\% variance & 64.86 & 72.92 & 77.55 & 80.10 & 81.54 \\
PCA 70\% variance & 63.24 & 71.45 & 76.19 & 78.92 & 80.38 \\
PCA 65\% variance & 61.81 & 70.16 & 75.18 & 77.94 & 79.46 \\
PCA 60\% variance & 59.68 & 68.26 & 73.54 & 76.41 & 77.99 \\
PCA 55\% variance & 56.85 & 65.34 & 70.70 & 73.65 & 75.24 \\
PCA 50\% variance & 53.53 & 61.97 & 67.66 & 70.88 & 72.65 \\
\bottomrule
\end{tabular}
\end{table}

\subsection{\rt{Sensitivity analysis for the hyper-parameter $\beta$ and the variance threshold}}
\label{app:persens}

\rt{Following Appendices~\ref{app:beta} and \ref{app:pca-variance}, DEFTS uses a fixed hyperparameter setting for all datasets ($\beta=0.1$ and a 95\% variance threshold) rather than tuning them individually. In this section, we evaluate the sensitivity of DEFTS to these hyperparameter choices by considering two experiments. First, we perform oracle tuning, where $\beta$ and the threshold are optimized directly on each dataset's own test accuracy. Second, we use a leave-one-dataset-out approach: for any given dataset, we tune the hyperparameters on the other 11 datasets before applying them to the held-out one. As in Appendices~\ref{app:beta} and \ref{app:pca-variance}, the hyperparameters are searched over 12 values for $\beta\in\{0.02, 0.05, 0.1, 0.2, \dots, 1.0\}$ and thresholds in $\{80\%, 90\%, 95\%, 99\%, 99.9\%\}$. Table~\ref{tab:persens_summary} reports the average change in accuracy relative to the fixed setting, while Tables~\ref{tab:persens_regret} and \ref{tab:persens_lodo} provide detailed breakdowns for each dataset.}

\rt{The results indicate that oracle tuning yields an average improvement of less than one accuracy point when $k > 1$. Furthermore, tuning on the other datasets alters the accuracy by at most $0.14$ points on average, demonstrating that the fixed setting transfers robustly. The 1-shot setting remains the most sensitive case, as also indicated by Figure~\ref{fig:beta}.}

\begin{table}[!ht]
\centering
\caption{\rt{Average accuracy differences compared to the fixed setting with $\beta=0.1$, 95\% threshold.The results are obtained using model backbone ViT-L/14 and are averaged over 10 random seeds. First row: oracle tuning on its own dataset. Second row: leave-one-out tuning on the other 11 datasets.}}
\label{tab:persens_summary}
\rt{\small
\renewcommand{\arraystretch}{1.3}
\begin{tabular}{@{}p{0.40\linewidth} ccccc@{}}
\toprule
\textbf{Change relative to the fixed setting (points, average over 12 datasets)} & \textbf{1 shot} & \textbf{2 shots} & \textbf{4 shots} & \textbf{8 shots} & \textbf{16 shots} \\
\midrule
Tuned for each dataset alone (oracle) & $+$1.82 & $+$0.84 & $+$0.61 & $+$0.33 & $+$0.31 \\
Tuned on the other 11 datasets & $-$0.60 & $-$0.14 & $+$0.01 & $+$0.02 & $+$0.06 \\
\bottomrule
\end{tabular}}
\end{table}

\begin{table}[ht]
\centering
\caption{\rt{Accuracy differences of the oracle tuning that picks $\beta$ and the variance threshold separately for each dataset over the fixed default ($\beta=0.1$, 95\%).
The results are obtained using model backbone ViT-L/14 and are averaged over 10 random seeds.}}
\label{tab:persens_regret}
\rt{\small
\renewcommand{\arraystretch}{1.1}
\begin{tabular}{@{}l ccccc@{}}
\toprule
\textbf{Dataset} & $k=1$ & $k=2$ & $k=4$ & $k=8$ & $k=16$ \\
\midrule
CIFAR10 & 0.00 & 0.14 & 0.05 & 0.03 & 0.04 \\
CIFAR100 & 1.60 & 1.65 & 0.86 & 0.64 & 0.53 \\
Caltech101 & 1.24 & 0.64 & 0.44 & 0.29 & 0.31 \\
DTD & 0.77 & 1.01 & 1.60 & 0.21 & 0.05 \\
EuroSAT & 3.45 & 1.60 & 1.17 & 1.22 & 1.10 \\
Aircraft & 1.25 & 1.34 & 1.57 & 0.55 & 0.76 \\
Flowers102 & 2.11 & 0.87 & 0.31 & 0.16 & 0.19 \\
Food101 & 1.39 & 0.63 & 0.52 & 0.34 & 0.25 \\
ImageNet & 5.33 & 1.13 & 0.14 & 0.12 & 0.19 \\
Pets & 0.97 & 0.60 & 0.44 & 0.17 & 0.26 \\
SUN397 & 2.78 & 0.28 & 0.00 & 0.00 & 0.00 \\
Cars & 0.92 & 0.16 & 0.19 & 0.28 & 0.03 \\
\midrule
Average of the 12 datasets & 1.82 & 0.84 & 0.61 & 0.33 & 0.31 \\
Largest over datasets & 5.33 & 1.65 & 1.60 & 1.22 & 1.10 \\
\bottomrule
\end{tabular}}
\end{table}

\begin{table}[ht]
\centering
\caption{\rt{Accuracy differences of the leave-one-out tuning that picks $\beta$ and the variance threshold on the other 11 datasets for each dataset over the fixed default ($\beta=0.1$, 95\%). The results are obtained using model backbone ViT-L/14 and are averaged over 10 random seeds.}}
\label{tab:persens_lodo}
\rt{\small
\renewcommand{\arraystretch}{1.1}
\begin{tabular}{@{}l ccccc@{}}
\toprule
\textbf{Dataset} & $k=1$ & $k=2$ & $k=4$ & $k=8$ & $k=16$ \\
\midrule
CIFAR10 & $-$0.05 & $+$0.09 & $+$0.03 & $+$0.03 & $+$0.02 \\
CIFAR100 & $+$1.59 & $+$0.22 & $+$0.55 & $+$0.55 & $+$0.49 \\
Caltech101 & $+$0.55 & $+$0.44 & $+$0.28 & $+$0.25 & $+$0.22 \\
DTD & $-$0.80 & $+$0.08 & $-$0.13 & $-$0.16 & $-$0.10 \\
EuroSAT & $-$1.19 & $+$0.79 & $+$0.77 & $+$1.09 & $+$1.10 \\
Aircraft & $-$0.91 & $-$0.43 & $-$0.44 & $-$0.27 & $-$0.05 \\
Flowers102 & $-$3.78 & $+$0.75 & $+$0.21 & $+$0.07 & $+$0.07 \\
Food101 & $+$0.75 & $-$0.45 & $-$0.25 & $-$0.13 & $-$0.03 \\
ImageNet & $-$2.92 & $-$1.77 & $-$0.86 & $-$0.91 & $-$0.89 \\
Pets & $+$0.59 & $+$0.18 & $+$0.44 & $+$0.10 & $+$0.23 \\
SUN397 & $-$0.66 & $-$0.66 & $-$0.21 & $-$0.46 & $-$0.36 \\
Cars & $-$0.36 & $-$0.97 & $-$0.26 & $+$0.11 & $+$0.03 \\
\midrule
Average of the 12 datasets & $-$0.60 & $-$0.14 & $+$0.01 & $+$0.02 & $+$0.06 \\
Largest gain over datasets & $+$1.59 & $+$0.79 & $+$0.77 & $+$1.09 & $+$1.10 \\
Largest loss over datasets & $-$3.78 & $-$1.77 & $-$0.86 & $-$0.91 & $-$0.89 \\
\bottomrule
\end{tabular}}
\end{table}

\subsection{Detailed experiment results with different number of shots}
\label{app:detailedresults}

In this section, we perform the same simulations for Table~\ref{tab:ensemble_improvements_datasets_k1}
with different number of shots $k \in \{1,4,8,16\}$. The results are presented in 
Table~\ref{tab:ensemble_improvements_datasets_1} to ~\ref{tab:ensemble_improvements_datasets_4}.

\begin{table*}[htbp]
\centering
\caption{Number of shots $k = 1$. Classification accuracy (\%) and performance improvements across datasets. The values in parentheses denote the absolute performance change after ensembling with DEFTS and the $90\%$ confidence interval. Positive gains are highlighted in green, and negative changes in red.}
\label{tab:ensemble_improvements_datasets_k1}

\newcommand{\gain}[3]{\begin{tabular}{@{}c@{}}#1 \\[-1ex] \textcolor{green!60!black}{\tiny (+#2 $\pm$#3)}\end{tabular}}
\newcommand{\loss}[3]{\begin{tabular}{@{}c@{}}#1 \\[-1ex] \textcolor{red!80!black}{\tiny (-#2 $\pm$#3)}\end{tabular}}

\renewcommand{\arraystretch}{1.0} 
\setlength{\tabcolsep}{4.5pt}

\resizebox{\textwidth}{!}{
\scriptsize 
\begin{tabular}{@{}l ccccccccc@{}}
\toprule
\textbf{Dataset} & \textbf{ZS} & \textbf{SimpleShot} & \textbf{Linear Probe} & \textbf{Tip-Adapter} & \textbf{APE} & \textbf{TIP-X} & \textbf{TAMP+ZS} & \textbf{LDA+ZS} & \textbf{DEFTS} \\
\midrule
C10 & 
\gain{94.89}{0.21}{0.17} & \gain{60.57}{30.13}{2.99} & \gain{70.54}{23.49}{4.18} & \gain{95.01}{0.12}{0.14} & \gain{95.05}{0.10}{0.14} & \gain{95.03}{0.11}{0.13} & \gain{94.81}{0.26}{0.14} & \gain{94.90}{0.21}{0.15} & 95.14 \\
\addlinespace[2pt]
C100 & 
\gain{77.27}{1.27}{0.18} & \gain{33.04}{29.25}{1.04} & \gain{38.88}{26.85}{1.05} & \gain{77.41}{1.17}{0.16} & \gain{77.85}{0.79}{0.11} & \gain{77.58}{0.99}{0.15} & \gain{77.67}{0.60}{0.12} & \gain{77.29}{1.27}{0.19} & 77.28 \\
\addlinespace[2pt]
Cal101 & 
\gain{94.62}{0.86}{0.27} & \gain{79.69}{11.04}{1.14} & \gain{85.78}{7.03}{0.84} & \gain{94.88}{0.62}{0.24} & \gain{95.30}{0.24}{0.27} & \gain{94.99}{0.51}{0.23} & \loss{95.37}{0.01}{0.19} & \gain{94.65}{0.83}{0.27} & 94.73 \\
\addlinespace[2pt]
DTD & 
\gain{53.03}{6.50}{0.63} & \gain{34.44}{16.09}{0.92} & \gain{38.20}{14.78}{1.06} & \gain{53.45}{6.17}{0.52} & \gain{54.08}{5.73}{0.58} & \gain{53.99}{5.60}{0.58} & \gain{56.69}{2.73}{0.49} & \gain{53.04}{6.48}{0.62} & 60.31 \\
\addlinespace[2pt]
Euro & 
\gain{61.17}{2.13}{0.47} & \gain{60.79}{11.39}{1.77} & \gain{60.83}{11.91}{2.01} & \gain{62.40}{1.35}{0.43} & \gain{64.31}{0.19}{0.44} & \gain{63.03}{0.94}{0.43} & \gain{60.53}{2.25}{0.43} & \gain{61.31}{2.06}{0.48} & 64.62 \\
\addlinespace[2pt]
Air. & 
\gain{32.18}{7.16}{0.60} & \gain{25.63}{9.27}{0.68} & \gain{26.85}{8.92}{0.62} & \gain{33.28}{6.17}{0.60} & \gain{34.85}{4.79}{0.83} & \gain{33.71}{5.80}{0.63} & \gain{37.09}{2.36}{0.37} & \gain{32.18}{7.16}{0.59} & 40.45 \\
\addlinespace[2pt]
Flow. & 
\gain{78.40}{9.38}{0.36} & \gain{82.19}{8.85}{0.89} & \gain{84.71}{7.20}{0.55} & \gain{79.11}{8.77}{0.33} & \gain{81.01}{7.26}{0.37} & \gain{79.53}{8.45}{0.33} & \gain{84.87}{3.50}{0.27} & \gain{78.52}{9.27}{0.36} & 91.19 \\
\addlinespace[2pt]
Food & 
\gain{93.11}{0.09}{0.10} & \gain{59.40}{23.15}{0.47} & \gain{64.31}{21.04}{0.43} & \gain{93.12}{0.05}{0.10} & \loss{93.18}{0.04}{0.11} & \gain{93.14}{0.02}{0.10} & \loss{93.19}{0.18}{0.08} & \gain{93.11}{0.07}{0.10} & 92.01 \\
\addlinespace[2pt]
INet & 
\gain{74.35}{1.54}{0.13} & \gain{40.94}{16.19}{0.29} & \gain{44.25}{14.78}{0.16} & \gain{74.56}{1.27}{0.13} & \gain{75.11}{0.62}{0.13} & \gain{74.68}{1.12}{0.12} & \loss{74.43}{0.70}{0.09} & \gain{74.41}{1.46}{0.13} & 70.92 \\
\addlinespace[2pt]
Pet & 
\gain{92.82}{0.58}{0.30} & \gain{49.77}{31.63}{1.10} & \gain{57.06}{29.04}{1.25} & \gain{92.97}{0.44}{0.27} & \gain{93.14}{0.28}{0.31} & \gain{93.01}{0.40}{0.28} & \gain{92.97}{0.28}{0.26} & \gain{92.86}{0.54}{0.30} & 92.85 \\
\addlinespace[2pt]
SUN & 
\gain{67.35}{3.64}{0.28} & \gain{40.37}{14.28}{0.23} & \gain{44.74}{12.95}{0.28} & \gain{67.64}{3.35}{0.31} & \gain{68.23}{2.68}{0.30} & \gain{67.90}{3.09}{0.31} & \gain{69.67}{0.13}{0.16} & \gain{67.38}{3.60}{0.27} & 67.86 \\
\addlinespace[2pt]
Cars & 
\gain{76.61}{3.05}{0.43} & \gain{43.79}{19.20}{0.66} & \gain{47.07}{19.19}{0.40} & \gain{76.80}{2.85}{0.36} & \gain{77.37}{2.26}{0.39} & \gain{76.92}{2.76}{0.37} & \gain{77.29}{1.02}{0.30} & \gain{76.65}{3.01}{0.42} & 78.10 \\
\midrule
AVG & 
\gain{\textbf{74.65}}{3.03}{0.46} & \gain{\textbf{50.88}}{18.37}{1.25} & \gain{\textbf{55.27}}{16.43}{1.15} & \gain{\textbf{75.05}}{2.69}{0.43} & \gain{\textbf{75.79}}{2.07}{0.38} & \gain{\textbf{75.29}}{2.48}{0.41} & \gain{\textbf{76.21}}{1.02}{0.21} & \gain{\textbf{74.69}}{3.00}{0.45} & \textbf{77.12} \\
\bottomrule
\end{tabular}
}
\end{table*}

\begin{table*}[htbp]
\centering
\caption{Number of shots $k=4$. Classification accuracy (\%) and performance improvements across datasets. The values in parentheses denote the absolute performance change after ensembling with DEFTS and the $90\%$ confidence interval. Positive gains are highlighted in green, and negative changes in red.}
\label{tab:ensemble_improvements_datasets_2}

\newcommand{\gain}[3]{\begin{tabular}{@{}c@{}}#1 \\[-1ex] \textcolor{green!60!black}{\tiny (+#2 $\pm$#3)}\end{tabular}}
\newcommand{\loss}[3]{\begin{tabular}{@{}c@{}}#1 \\[-1ex] \textcolor{red!80!black}{\tiny (-#2 $\pm$#3)}\end{tabular}}

\renewcommand{\arraystretch}{1.0} 
\setlength{\tabcolsep}{4.5pt}

\resizebox{\textwidth}{!}{
\scriptsize 
\begin{tabular}{@{}l ccccccccc@{}}
\toprule
\textbf{Dataset} & \textbf{ZS} & \textbf{SimpleShot} & \textbf{Linear Probe} & \textbf{Tip-Adapter} & \textbf{APE} & \textbf{TIP-X} & \textbf{TAMP+ZS} & \textbf{LDA+ZS} & \textbf{DEFTS} \\
\midrule
C10 & 
\gain{94.89}{0.66}{0.12} & \gain{87.68}{7.94}{1.13} & \gain{89.68}{5.85}{0.63} & \gain{95.05}{0.32}{0.26} & \gain{94.22}{0.36}{0.61} & \gain{94.34}{0.45}{0.15} & \gain{94.82}{0.64}{0.24} & \gain{94.81}{0.24}{0.25} & 95.56 \\
\addlinespace[2pt]
C100 & 
\gain{77.27}{1.83}{0.51} & \gain{63.50}{15.21}{0.49} & \gain{69.05}{9.51}{0.67} & \gain{77.96}{0.77}{0.49} & \gain{78.27}{1.11}{0.55} & \gain{69.32}{0.96}{0.92} & \gain{78.08}{1.12}{0.46} & \gain{79.66}{0.55}{0.76} & 78.60 \\
\addlinespace[2pt]
Cal101 & 
\gain{94.62}{1.64}{0.52} & \gain{94.16}{2.52}{0.27} & \gain{94.78}{1.67}{0.25} & \gain{94.99}{1.35}{0.47} & \gain{95.00}{1.33}{0.47} & \gain{95.12}{0.15}{0.19} & \gain{95.42}{0.80}{0.51} & \gain{97.08}{0.27}{0.45} & 96.55 \\
\addlinespace[2pt]
DTD & 
\gain{53.03}{12.93}{1.02} & \gain{57.94}{9.92}{0.96} & \gain{60.61}{7.38}{0.96} & \gain{60.56}{5.81}{2.12} & \gain{64.12}{2.50}{1.53} & \gain{61.36}{0.00}{0.00} & \gain{60.88}{4.91}{1.07} & \gain{66.33}{1.80}{1.68} & 66.86 \\
\addlinespace[2pt]
Euro & 
\gain{61.17}{9.87}{0.67} & \gain{76.66}{2.16}{2.37} & \loss{78.53}{0.57}{3.00} & \loss{75.48}{0.55}{1.83} & \gain{77.20}{0.29}{2.53} & \gain{67.21}{0.25}{0.28} & \gain{60.57}{10.41}{1.11} & \loss{82.42}{0.43}{1.37} & 71.49 \\
\addlinespace[2pt]
Air. & 
\gain{32.18}{14.15}{1.34} & \gain{41.84}{7.15}{0.88} & \gain{40.61}{8.51}{1.78} & \gain{39.60}{6.85}{2.16} & \gain{40.57}{3.95}{1.65} & \gain{45.59}{1.86}{0.66} & \gain{43.20}{1.74}{1.12} & \gain{46.20}{1.33}{1.56} & 49.00 \\
\addlinespace[2pt]
Flow. & 
\gain{78.40}{18.80}{0.25} & \gain{96.06}{1.33}{0.17} & \gain{96.65}{0.24}{0.64} & \gain{94.96}{1.90}{0.57} & \gain{96.15}{0.85}{0.25} & \gain{94.19}{1.05}{0.18} & \gain{94.11}{3.00}{0.39} & \gain{97.80}{0.09}{0.05} & 97.22 \\
\addlinespace[2pt]
Food & 
\gain{93.11}{0.51}{0.07} & \gain{85.61}{7.18}{0.33} & \gain{87.33}{5.53}{0.55} & \gain{93.34}{0.16}{0.11} & \gain{93.30}{0.11}{0.11} & \gain{89.59}{0.42}{0.33} & \gain{93.37}{0.24}{0.13} & \gain{93.66}{0.09}{0.13} & 92.96 \\
\addlinespace[2pt]
INet & 
\gain{74.35}{2.54}{0.47} & \gain{66.04}{11.34}{0.34} & \gain{69.12}{8.56}{0.26} & \gain{75.39}{0.00}{0.00} & \gain{75.84}{0.00}{0.00} & \gain{67.15}{0.00}{0.00} & \gain{78.31}{0.02}{0.04} & \gain{78.46}{0.00}{0.00} & 77.71 \\
\addlinespace[2pt]
Pet & 
\gain{92.82}{1.80}{0.40} & \gain{80.84}{13.60}{0.79} & \gain{82.02}{12.05}{1.56} & \gain{93.25}{1.35}{0.40} & \gain{93.20}{1.50}{0.34} & \gain{93.60}{0.56}{0.20} & \gain{92.86}{1.83}{0.42} & \gain{93.53}{1.16}{0.48} & 94.54 \\
\addlinespace[2pt]
SUN & 
\gain{67.35}{5.17}{0.53} & \gain{65.96}{8.65}{0.27} & \gain{67.94}{6.67}{0.36} & \gain{70.34}{3.32}{0.80} & \gain{70.88}{2.61}{1.12} & \gain{66.02}{0.00}{0.00} & \loss{73.70}{0.00}{0.01} & \gain{74.68}{0.22}{0.21} & 74.77 \\
\addlinespace[2pt]
Cars & 
\gain{76.61}{6.09}{0.97} & \gain{69.67}{14.78}{0.62} & \gain{73.73}{10.81}{0.33} & \gain{77.71}{4.68}{1.03} & \gain{78.63}{5.01}{0.72} & \gain{79.65}{0.59}{0.58} & \gain{80.30}{1.67}{0.56} & \gain{80.81}{2.80}{0.96} & 84.65 \\
\midrule
AVG & 
\gain{\textbf{74.65}}{6.33}{0.91} & \gain{\textbf{73.83}}{8.48}{0.73} & \gain{\textbf{75.84}}{6.35}{0.66} & \gain{\textbf{79.05}}{2.16}{0.45} & \gain{\textbf{79.78}}{1.64}{0.36} & \gain{\textbf{76.93}}{0.52}{0.13} & \gain{\textbf{78.80}}{2.20}{0.46} & \gain{\textbf{82.12}}{0.68}{0.26} & \textbf{81.66} \\
\bottomrule
\end{tabular}
}
\end{table*}

\begin{table*}[htbp]
\centering
\caption{Number of shots $k=8$. Classification accuracy (\%) and performance improvements across datasets. The values in parentheses denote the absolute performance change after ensembling with DEFTS and the $90\%$ confidence interval. Positive gains are highlighted in green, and negative changes in red.}
\label{tab:ensemble_improvements_datasets_3}

\newcommand{\gain}[3]{\begin{tabular}{@{}c@{}}#1 \\[-1ex] \textcolor{green!60!black}{\tiny (+#2 $\pm$#3)}\end{tabular}}
\newcommand{\loss}[3]{\begin{tabular}{@{}c@{}}#1 \\[-1ex] \textcolor{red!80!black}{\tiny (-#2 $\pm$#3)}\end{tabular}}

\renewcommand{\arraystretch}{1.0} 
\setlength{\tabcolsep}{4.5pt}

\resizebox{\textwidth}{!}{
\scriptsize 
\begin{tabular}{@{}l ccccccccc@{}}
\toprule
\textbf{Dataset} & \textbf{ZS} & \textbf{SimpleShot} & \textbf{Linear Probe} & \textbf{Tip-Adapter} & \textbf{APE} & \textbf{TIP-X} & \textbf{TAMP+ZS} & \textbf{LDA+ZS} & \textbf{DEFTS} \\
\midrule
C10 & 
\gain{94.89}{0.59}{0.18} & \gain{92.41}{2.75}{0.68} & \gain{93.42}{2.13}{1.41} & \gain{95.17}{0.23}{0.11} & \gain{95.39}{0.20}{0.11} & \gain{94.64}{0.07}{0.05} & \gain{94.97}{0.53}{0.16} & \gain{95.81}{0.12}{0.09} & 95.60 \\
\addlinespace[2pt]
C100 & 
\gain{77.27}{3.21}{0.19} & \gain{70.57}{9.78}{0.60} & \gain{75.39}{4.52}{0.53} & \gain{78.19}{2.51}{0.22} & \gain{79.60}{1.16}{0.46} & \gain{70.81}{0.00}{0.00} & \gain{78.54}{1.91}{0.41} & \gain{81.87}{0.10}{0.27} & 80.25 \\
\addlinespace[2pt]
Cal101 & 
\gain{94.62}{2.86}{0.16} & \gain{95.53}{1.56}{0.29} & \gain{96.26}{0.45}{0.29} & \gain{95.90}{1.51}{0.42} & \gain{96.20}{1.23}{0.41} & \gain{95.83}{0.00}{0.00} & \gain{96.45}{0.92}{0.22} & \gain{97.61}{0.03}{0.09} & 97.08 \\
\addlinespace[2pt]
DTD & 
\gain{53.03}{17.40}{0.44} & \gain{64.58}{6.73}{0.87} & \gain{67.48}{2.91}{1.09} & \gain{66.25}{4.47}{0.96} & \gain{68.69}{2.96}{1.06} & \gain{63.94}{0.00}{0.00} & \gain{62.90}{7.17}{0.49} & \gain{71.49}{1.39}{0.65} & 70.56 \\
\addlinespace[2pt]
Euro & 
\gain{61.17}{12.86}{0.82} & \gain{80.45}{3.30}{0.34} & \loss{85.12}{1.21}{1.49} & \gain{80.26}{1.54}{1.11} & \gain{83.52}{0.56}{0.50} & \gain{69.18}{0.19}{0.36} & \gain{60.61}{12.58}{1.00} & \gain{87.30}{0.43}{0.56} & 74.35 \\
\addlinespace[2pt]
Air. & 
\gain{32.18}{21.41}{0.55} & \gain{47.96}{6.46}{0.47} & \gain{49.27}{4.46}{0.66} & \gain{46.50}{6.65}{1.25} & \gain{43.00}{8.34}{1.33} & \gain{47.93}{0.00}{0.00} & \gain{47.65}{5.36}{0.94} & \gain{55.29}{2.25}{0.64} & 53.98 \\
\addlinespace[2pt]
Flow. & 
\gain{78.40}{19.66}{0.20} & \gain{97.48}{0.61}{0.11} & \loss{98.19}{0.46}{0.22} & \gain{96.77}{1.06}{0.14} & \gain{97.59}{0.46}{0.18} & \gain{94.62}{0.14}{0.18} & \gain{95.04}{2.92}{0.31} & \gain{98.66}{0.12}{0.07} & 98.06 \\
\addlinespace[2pt]
Food & 
\gain{93.11}{0.54}{0.10} & \gain{89.05}{4.10}{0.16} & \gain{90.39}{2.54}{0.32} & \gain{93.43}{0.25}{0.14} & \gain{93.51}{0.27}{0.18} & \gain{90.28}{0.00}{0.00} & \gain{93.59}{0.03}{0.09} & \loss{93.93}{0.10}{0.12} & 93.15 \\
\addlinespace[2pt]
INet & 
\gain{74.35}{6.31}{0.09} & \gain{72.64}{7.20}{0.12} & \gain{75.74}{4.19}{0.13} & \gain{76.94}{3.71}{0.21} & \gain{77.73}{2.99}{0.32} & \loss{74.10}{0.34}{0.70} & \gain{79.54}{0.61}{0.20} & \gain{81.10}{0.13}{0.08} & 79.93 \\
\addlinespace[2pt]
Pet & 
\gain{92.82}{2.26}{0.24} & \gain{86.71}{8.39}{0.71} & \gain{90.02}{4.52}{0.96} & \gain{93.61}{1.44}{0.38} & \gain{93.83}{1.33}{0.41} & \gain{93.51}{0.31}{0.12} & \gain{93.01}{2.11}{0.27} & \gain{95.24}{0.18}{0.29} & 95.25 \\
\addlinespace[2pt]
SUN & 
\gain{67.35}{9.75}{0.14} & \gain{72.08}{5.19}{0.25} & \gain{74.04}{2.93}{0.37} & \gain{73.32}{3.83}{0.24} & \gain{74.37}{2.98}{0.29} & \gain{70.01}{0.00}{0.00} & \gain{75.71}{1.11}{0.29} & \gain{78.27}{0.22}{0.13} & 77.27 \\
\addlinespace[2pt]
Cars & 
\gain{76.61}{9.98}{0.20} & \gain{76.12}{10.39}{0.37} & \gain{81.96}{4.30}{0.75} & \gain{80.04}{6.24}{0.38} & \gain{81.23}{4.96}{0.37} & \gain{81.22}{0.00}{0.00} & \gain{82.03}{4.38}{0.42} & \gain{86.28}{1.01}{0.22} & 86.66 \\
\midrule
AVG & 
\gain{\textbf{74.65}}{8.90}{1.10} & \gain{\textbf{78.80}}{5.54}{0.47} & \gain{\textbf{81.44}}{2.61}{0.36} & \gain{\textbf{81.37}}{2.79}{0.35} & \gain{\textbf{82.06}}{2.29}{0.38} & \gain{\textbf{78.84}}{0.03}{0.06} & \gain{\textbf{80.00}}{3.30}{0.54} & \gain{\textbf{85.24}}{0.49}{0.13} & \textbf{83.51} \\
\bottomrule
\end{tabular}
}
\end{table*}

\begin{table*}[htbp]
\centering
\caption{Number of shots $k=16$. Classification accuracy (\%) and performance improvements across datasets. The values in parentheses denote the absolute performance change after ensembling with DEFTS and the $90\%$ confidence interval. Positive gains are highlighted in green, and negative changes in red.}
\label{tab:ensemble_improvements_datasets_4}

\newcommand{\gain}[3]{\begin{tabular}{@{}c@{}}#1 \\[-1ex] \textcolor{green!60!black}{\tiny (+#2 $\pm$#3)}\end{tabular}}
\newcommand{\loss}[3]{\begin{tabular}{@{}c@{}}#1 \\[-1ex] \textcolor{red!80!black}{\tiny (-#2 $\pm$#3)}\end{tabular}}

\renewcommand{\arraystretch}{1.0} 
\setlength{\tabcolsep}{4.5pt}

\resizebox{\textwidth}{!}{
\scriptsize 
\begin{tabular}{@{}l ccccccccc@{}}
\toprule
\textbf{Dataset} & \textbf{ZS} & \textbf{SimpleShot} & \textbf{Linear Probe} & \textbf{Tip-Adapter} & \textbf{APE} & \textbf{TIP-X} & \textbf{TAMP+ZS} & \textbf{LDA+ZS} & \textbf{DEFTS} \\
\midrule
C10 & 
\gain{94.89}{0.70}{0.11} & \gain{94.75}{1.20}{0.42} & \gain{95.35}{0.66}{0.47} & \gain{95.52}{0.16}{0.11} & \gain{95.76}{0.08}{0.16} & \gain{95.26}{0.00}{0.00} & \gain{95.07}{0.53}{0.12} & \gain{96.04}{0.23}{0.18} & 95.67 \\
\addlinespace[2pt]
C100 & 
\gain{77.27}{4.28}{0.13} & \gain{74.89}{6.73}{0.43} & \gain{79.65}{1.60}{0.42} & \gain{79.13}{2.46}{0.15} & \gain{80.40}{1.45}{0.28} & \gain{73.08}{0.00}{0.00} & \gain{78.95}{2.60}{0.20} & \gain{82.93}{0.09}{0.19} & 81.41 \\
\addlinespace[2pt]
Cal101 & 
\gain{94.62}{2.89}{0.10} & \gain{95.99}{1.48}{0.24} & \gain{97.09}{0.27}{0.08} & \gain{96.41}{1.21}{0.28} & \gain{96.87}{0.75}{0.21} & \gain{96.08}{0.20}{0.19} & \gain{96.64}{0.90}{0.16} & \loss{97.95}{0.01}{0.12} & 97.36 \\
\addlinespace[2pt]
DTD & 
\gain{53.03}{19.93}{0.44} & \gain{69.00}{5.25}{0.45} & \loss{74.12}{0.75}{0.84} & \gain{69.44}{5.04}{0.60} & \gain{72.93}{2.23}{0.62} & \loss{67.14}{0.05}{0.28} & \gain{65.07}{7.76}{0.74} & \gain{75.43}{1.08}{0.81} & 72.95 \\
\addlinespace[2pt]
Euro & 
\gain{61.17}{13.93}{0.60} & \gain{81.39}{2.49}{0.33} & \loss{89.47}{3.98}{1.44} & \gain{82.39}{0.74}{0.86} & \gain{86.81}{0.65}{0.30} & \gain{72.78}{0.58}{1.66} & \gain{60.73}{14.40}{0.83} & \loss{91.19}{0.19}{0.27} & 75.44 \\
\addlinespace[2pt]
Air. & 
\gain{32.18}{24.43}{0.53} & \gain{52.40}{5.47}{0.57} & \gain{56.20}{0.97}{0.34} & \gain{51.25}{6.05}{0.54} & \gain{47.66}{8.30}{0.82} & \gain{49.58}{0.00}{0.00} & \gain{51.19}{5.59}{0.81} & \gain{60.15}{2.87}{0.53} & 56.62 \\
\addlinespace[2pt]
Flow. & 
\gain{78.40}{20.11}{0.08} & \gain{98.25}{0.40}{0.11} & \loss{99.00}{0.49}{0.08} & \gain{97.38}{1.07}{0.12} & \gain{98.46}{0.30}{0.09} & \gain{95.39}{0.47}{0.50} & \gain{95.65}{2.89}{0.26} & \gain{99.03}{0.11}{0.06} & 98.49 \\
\addlinespace[2pt]
Food & 
\gain{93.11}{0.79}{0.09} & \gain{90.84}{2.43}{0.13} & \gain{92.23}{0.88}{0.19} & \gain{93.58}{0.34}{0.10} & \gain{93.76}{0.15}{0.11} & \gain{91.02}{0.00}{0.00} & \gain{93.60}{0.24}{0.14} & \loss{94.18}{0.02}{0.08} & 93.33 \\
\addlinespace[2pt]
INet & 
\gain{74.35}{7.60}{0.06} & \gain{76.08}{5.31}{0.13} & \gain{80.14}{1.27}{0.09} & \gain{78.37}{3.63}{0.27} & \gain{79.70}{2.41}{0.18} & \gain{76.73}{3.22}{0.84} & \gain{80.17}{1.75}{0.10} & \gain{80.38}{1.96}{0.12} & 81.41 \\
\addlinespace[2pt]
Pet & 
\gain{92.82}{2.77}{0.08} & \gain{89.98}{5.46}{0.42} & \gain{93.12}{2.15}{0.26} & \gain{93.72}{1.79}{0.35} & \gain{94.14}{1.42}{0.31} & \gain{93.76}{0.02}{0.05} & \gain{93.03}{2.39}{0.25} & \gain{95.86}{0.04}{0.16} & 95.52 \\
\addlinespace[2pt]
SUN & 
\gain{67.35}{11.59}{0.16} & \gain{75.37}{3.28}{0.16} & \gain{78.06}{0.67}{0.15} & \gain{74.99}{4.04}{0.19} & \gain{76.27}{3.06}{0.23} & \gain{73.73}{0.46}{0.80} & \gain{77.21}{1.67}{0.18} & \gain{78.85}{1.49}{0.15} & 78.73 \\
\addlinespace[2pt]
Cars & 
\gain{76.61}{11.47}{0.11} & \gain{79.73}{8.53}{0.34} & \gain{87.02}{1.18}{0.34} & \gain{82.06}{5.77}{0.29} & \gain{83.34}{4.47}{0.34} & \gain{82.45}{0.24}{0.35} & \gain{83.36}{4.71}{0.40} & \gain{88.35}{0.73}{0.12} & 88.21 \\
\midrule
AVG & 
\gain{\textbf{74.65}}{10.04}{1.20} & \gain{\textbf{81.56}}{4.00}{0.37} & \gain{\textbf{85.12}}{0.37}{0.27} & \gain{\textbf{82.85}}{2.69}{0.33} & \gain{\textbf{83.84}}{2.11}{0.35} & \gain{\textbf{80.58}}{0.43}{0.20} & \gain{\textbf{80.89}}{3.78}{0.59} & \gain{\textbf{86.70}}{0.70}{0.16} & \textbf{84.59} \\
\bottomrule
\end{tabular}
}
\end{table*}

\subsection{Experiments with different model backbone}
\label{app:backbones}

In this section, we evaluate the performances of DEFTS and 
other baselines with different model backbone considered in the main paper. 
To be specific, we consider ViT-B/32 and ViT-H-14, different than ViT-L/14 used
Section~\ref{sec:experiments}. The experiments in this section are performed exactly as the one for Table~\ref{tab:ensemble_improvements_k_shots}, with only model backbone changed. Results
shown in Table~\ref{tab:ensemble_improvements_combined} demonstrate that the DEFTS improvements over other competing methods are consistent across different model backbones.

\begin{table*}[htbp]
\centering
\caption{Different model backbones. Average classification accuracy (\%) and performance improvements across 12 datasets for different numbers of shots $k$. Results are reported for two distinct model architectures: \textbf{ViT-B/32} and \textbf{ViT-H/14}. The values in parentheses denote the absolute performance change after ensembling with DEFTS and the $90\%$ confidence interval. Positive gains are highlighted in green, and negative changes in red. The best baseline accuracy for each $k$-shot setting (before adding DEFTS) is highlighted in \textbf{bold}.}
\label{tab:ensemble_improvements_combined}

\newcommand{\gain}[3]{\begin{tabular}{@{}c@{}}#1 \\[-0.5ex] \textcolor{green!60!black}{\scriptsize (+#2 $\pm$#3)}\end{tabular}}
\newcommand{\loss}[3]{\begin{tabular}{@{}c@{}}#1 \\[-0.5ex] \textcolor{red!80!black}{\scriptsize (-#2 $\pm$#3)}\end{tabular}}

\renewcommand{\arraystretch}{1.2}
\setlength{\tabcolsep}{4.5pt}

\resizebox{\textwidth}{!}{
\small 
\begin{tabular}{@{}l ccccccccc@{}}
\toprule
\textbf{$k$-shot} & \textbf{ZS+DEFTS} & \textbf{SimpleShot} & \textbf{Linear Probe} & \textbf{Tip-Adapter} & \textbf{APE} & \textbf{TIP-X} & \textbf{TAMP+ZS} & \textbf{LDA+ZS} & \textbf{DEFTS alone} \\
\midrule
\multicolumn{10}{c}{\textbf{Model Backbone: ViT-B/32 Backbone}} \\
\midrule
$k=1$ & 
\gain{64.56}{2.10}{0.48} & \gain{40.45}{16.26}{1.23} & \gain{42.21}{15.60}{1.25} & \gain{65.01}{1.72}{0.46} & \gain{65.78}{1.07}{0.42} & \gain{65.27}{1.47}{0.45} & \gain{64.92}{0.95}{0.36} & \gain{64.66}{2.01}{0.47} & \textbf{65.08} \\
\addlinespace
$k=2$ & 
\gain{64.56}{2.12}{0.58} & \gain{52.33}{16.12}{1.68} & \gain{52.56}{15.96}{1.50} & \gain{66.74}{0.44}{0.24} & \gain{67.66}{0.22}{0.17} & \gain{63.02}{1.77}{0.27} & \gain{65.73}{1.09}{0.43} & \gain{67.31}{0.39}{0.40} & \textbf{67.90} \\
\addlinespace
$k=4$ & 
\gain{64.56}{5.36}{0.92} & \gain{61.45}{10.25}{1.08} & \gain{63.77}{7.37}{1.10} & \gain{68.87}{1.59}{0.38} & \gain{69.67}{1.34}{0.39} & \gain{64.72}{0.54}{0.15} & \gain{68.00}{2.12}{0.51} & \gain{\textbf{72.13}}{0.76}{0.34} & 70.37 \\
\addlinespace
$k=8$ & 
\gain{64.56}{8.47}{1.09} & \gain{67.41}{6.88}{0.76} & \gain{71.21}{2.30}{0.82} & \gain{71.87}{2.27}{0.32} & \gain{73.06}{1.76}{0.29} & \gain{66.51}{0.05}{0.07} & \gain{69.32}{3.57}{0.70} & \gain{\textbf{76.24}}{0.47}{0.13} & 72.70 \\
\addlinespace
$k=16$ & 
\gain{64.56}{10.09}{1.20} & \gain{70.89}{5.02}{0.59} & \loss{76.04}{0.24}{0.54} & \gain{74.02}{2.03}{0.27} & \gain{75.34}{1.35}{0.23} & \gain{68.96}{0.12}{0.09} & \gain{70.41}{4.22}{0.73} & \gain{\textbf{77.27}}{1.42}{0.26} & 74.38 \\
\midrule
\multicolumn{10}{c}{\textbf{Model Backbone: ViT-H/14 Backbone}} \\
\midrule
$k=1$ & 
\gain{81.44}{1.87}{0.48} & \gain{60.81}{15.52}{0.92} & \gain{64.27}{14.11}{0.89} & \gain{81.70}{1.67}{0.46} & \gain{82.08}{1.33}{0.41} & \gain{81.83}{1.55}{0.44} & \gain{82.31}{0.66}{0.21} & \gain{81.48}{1.84}{0.47} & \textbf{82.61} \\
\addlinespace
$k=2$ & 
\gain{81.44}{1.64}{0.57} & \gain{73.52}{11.48}{1.13} & \gain{73.74}{11.18}{1.13} & \gain{83.27}{0.15}{0.09} & \gain{83.82}{0.11}{0.15} & \gain{80.88}{0.64}{0.16} & \gain{82.70}{0.59}{0.22} & \gain{83.12}{0.47}{0.21} & \textbf{84.65} \\
\addlinespace
$k=4$ & 
\gain{81.44}{3.78}{0.83} & \gain{81.42}{5.63}{0.53} & \gain{82.16}{4.70}{0.50} & \gain{85.01}{0.81}{0.28} & \gain{85.31}{0.73}{0.32} & \gain{82.86}{0.09}{0.04} & \gain{84.42}{1.06}{0.32} & \gain{\textbf{86.97}}{0.12}{0.16} & 85.95 \\
\addlinespace
$k=8$ & 
\gain{81.44}{5.90}{1.07} & \gain{85.39}{3.00}{0.31} & \gain{86.54}{1.77}{0.26} & \gain{86.67}{1.36}{0.30} & \gain{87.30}{1.02}{0.26} & \loss{84.88}{0.08}{0.11} & \gain{85.35}{1.94}{0.57} & \gain{\textbf{89.23}}{0.14}{0.09} & 87.39 \\
\addlinespace
$k=16$ & 
\gain{81.44}{6.82}{1.19} & \gain{87.47}{1.77}{0.19} & \gain{89.17}{0.12}{0.16} & \gain{87.83}{1.29}{0.27} & \gain{88.60}{0.90}{0.22} & \gain{86.89}{0.46}{0.16} & \gain{85.89}{2.33}{0.61} & \gain{\textbf{90.19}}{0.44}{0.12} & 88.21 \\
\bottomrule
\end{tabular}
}
\end{table*}

\subsection{Relative error reduction}
\label{app:normgain}

\rt{In Table~\ref{tab:ensemble_improvements_k_shots}, we report the absolute accuracy gains achieved by different methods when ensembled with DEFTS. However, absolute accuracy improvements can be misleading due to the ceiling effect, as strong baselines naturally have less room for improvement. To account for this, Table~\ref{tab:normgain} presents the relative error reduction, which quantifies the fraction of the base method's misclassifications that are successfully resolved by the ensemble.}

\begin{table}[!htbp]
\centering
\color{blue}
\caption{\rt{Average accuracy increase (\%) / relative error reduction by ensembling with DEFTS for different number of shots $k$. The average accuracy increase is the same as in Table~\ref{tab:ensemble_improvements_k_shots}.}}
\label{tab:normgain}
\small
\renewcommand{\arraystretch}{1.1}
\begin{tabular}{@{}l ccccc@{}}
\toprule
Base method & $k=1$ & $k=2$ & $k=4$ & $k=8$ & $k=16$ \\
\midrule
ZS & $+$3.03\,/\,12\% & $+$2.36\,/\,9\% & $+$6.33\,/\,25\% & $+$8.90\,/\,35\% & $+$10.04\,/\,40\% \\
SimpleShot & $+$18.37\,/\,37\% & $+$15.30\,/\,43\% & $+$8.48\,/\,32\% & $+$5.54\,/\,26\% & $+$4.00\,/\,22\% \\
Linear probe & $+$16.43\,/\,37\% & $+$14.07\,/\,41\% & $+$6.35\,/\,26\% & $+$2.61\,/\,14\% & $+$0.37\,/\,2\% \\
Tip-Adapter & $+$2.69\,/\,11\% & $+$0.76\,/\,3\% & $+$2.16\,/\,10\% & $+$2.79\,/\,15\% & $+$2.69\,/\,16\% \\
APE & $+$2.07\,/\,9\% & $+$0.41\,/\,2\% & $+$1.64\,/\,8\% & $+$2.29\,/\,13\% & $+$2.11\,/\,13\% \\
TIP-X & $+$2.48\,/\,10\% & $+$1.27\,/\,5\% & $+$0.52\,/\,2\% & $+$0.03\,/\,0\% & $+$0.43\,/\,2\% \\
TAMP+ZS & $+$1.02\,/\,4\% & $+$0.94\,/\,4\% & $+$2.20\,/\,10\% & $+$3.30\,/\,17\% & $+$3.78\,/\,20\% \\
LDA+ZS & $+$3.00\,/\,12\% & $+$1.12\,/\,5\% & $+$0.68\,/\,4\% & $+$0.49\,/\,3\% & $+$0.70\,/\,5\% \\
\bottomrule
\end{tabular}
\end{table}

\subsection{Ensembling comparison between DEFTS and TAMP}
\label{app:matched}

\rt{In this section, we compare DEFTS and TAMP further by comparing $X$ + DEFTS and $X$ + TAMP, where $X$ is an existing baseline such as TIP-X. Both ensembles are searched in the same way as described in Appendix~\ref{sec:hyperparametersearch}: the hyperparameters of $X$ and the weight of the added method are chosen jointly by leave-one-out accuracy on the support set. The weight of DEFTS and of TAMP takes the same values, $\{0, 0.5, 1.0, 1.5, 2.0, 2.5, 3.0, 3.5\}$ when $X$ is Tip-Adapter, TIP-X or APE, and 21 values in $[0,1]$ for the other base methods. TAMP has one more hyperparameter than DEFTS, its own mixing weight $\lambda$, so its search space is larger. The results are shown in Table~\ref{tab:matched_pooled}, which further demonstrates the advantage of DEFTS over TAMP.}

\begin{table}[!t]
\centering
\caption{\rt{Accuracy gain over the baseline X by ensembling with DEFTS or TAMP for each $k$. The results are obtained using model backbone ViT-L/14 and are averaged over 10 random seeds.}}
\label{tab:matched_pooled}
\small
\renewcommand{\arraystretch}{1.05}
\begin{tabular}{@{}l c ccc@{}}
\toprule
Base X & X & $+$DEFTS & $+$TAMP & DEFTS $-$ TAMP \\
\midrule
\multicolumn{5}{@{}l}{\emph{$k=1$}} \\
ZS & 74.65 & \textcolor{green!60!black}{$+$3.03\,{\tiny($\pm$0.46)}} & \textcolor{green!60!black}{$+$1.56\,{\tiny($\pm$0.34)}} & \textcolor{green!60!black}{$+$1.47\,{\tiny($\pm$0.18)}} \\
SimpleShot & 50.88 & \textcolor{green!60!black}{$+$18.37\,{\tiny($\pm$1.25)}} & \textcolor{green!60!black}{$+$11.96\,{\tiny($\pm$0.84)}} & \textcolor{green!60!black}{$+$6.41\,{\tiny($\pm$0.56)}} \\
Linear Probe & 55.27 & \textcolor{green!60!black}{$+$16.43\,{\tiny($\pm$1.15)}} & \textcolor{green!60!black}{$+$10.98\,{\tiny($\pm$0.80)}} & \textcolor{green!60!black}{$+$5.45\,{\tiny($\pm$0.48)}} \\
Tip-Adapter & 75.05 & \textcolor{green!60!black}{$+$2.69\,{\tiny($\pm$0.43)}} & \textcolor{green!60!black}{$+$1.25\,{\tiny($\pm$0.32)}} & \textcolor{green!60!black}{$+$1.44\,{\tiny($\pm$0.17)}} \\
TIP-X & 75.29 & \textcolor{green!60!black}{$+$2.48\,{\tiny($\pm$0.41)}} & \textcolor{green!60!black}{$+$1.09\,{\tiny($\pm$0.30)}} & \textcolor{green!60!black}{$+$1.39\,{\tiny($\pm$0.17)}} \\
APE & 75.79 & \textcolor{green!60!black}{$+$2.07\,{\tiny($\pm$0.38)}} & \textcolor{green!60!black}{$+$0.72\,{\tiny($\pm$0.27)}} & \textcolor{green!60!black}{$+$1.35\,{\tiny($\pm$0.17)}} \\
TAMP+ZS & 76.21 & \textcolor{green!60!black}{$+$1.02\,{\tiny($\pm$0.21)}} & -- & -- \\
LDA+ZS & 74.69 & \textcolor{green!60!black}{$+$3.00\,{\tiny($\pm$0.45)}} & \textcolor{green!60!black}{$+$1.54\,{\tiny($\pm$0.34)}} & \textcolor{green!60!black}{$+$1.46\,{\tiny($\pm$0.18)}} \\
\midrule
\multicolumn{5}{@{}l}{\emph{$k=2$}} \\
ZS & 74.65 & \textcolor{green!60!black}{$+$2.36\,{\tiny($\pm$0.69)}} & \textcolor{green!60!black}{$+$2.00\,{\tiny($\pm$0.58)}} & $+$0.36\,{\tiny($\pm$0.44)} \\
SimpleShot & 64.46 & \textcolor{green!60!black}{$+$15.30\,{\tiny($\pm$1.30)}} & \textcolor{green!60!black}{$+$13.08\,{\tiny($\pm$1.50)}} & \textcolor{green!60!black}{$+$2.22\,{\tiny($\pm$0.49)}} \\
Linear Probe & 65.74 & \textcolor{green!60!black}{$+$14.07\,{\tiny($\pm$1.07)}} & \textcolor{green!60!black}{$+$12.99\,{\tiny($\pm$1.15)}} & \textcolor{green!60!black}{$+$1.08\,{\tiny($\pm$0.37)}} \\
Tip-Adapter & 76.57 & \textcolor{green!60!black}{$+$0.76\,{\tiny($\pm$0.31)}} & \textcolor{green!60!black}{$+$0.90\,{\tiny($\pm$0.33)}} & $-$0.14\,{\tiny($\pm$0.30)} \\
TIP-X & 75.19 & \textcolor{green!60!black}{$+$1.27\,{\tiny($\pm$0.18)}} & \textcolor{green!60!black}{$+$0.99\,{\tiny($\pm$0.18)}} & \textcolor{green!60!black}{$+$0.28\,{\tiny($\pm$0.17)}} \\
APE & 77.34 & \textcolor{green!60!black}{$+$0.41\,{\tiny($\pm$0.25)}} & \textcolor{green!60!black}{$+$0.36\,{\tiny($\pm$0.27)}} & $+$0.04\,{\tiny($\pm$0.38)} \\
TAMP+ZS & 76.65 & \textcolor{green!60!black}{$+$0.94\,{\tiny($\pm$0.32)}} & -- & -- \\
LDA+ZS & 76.37 & \textcolor{green!60!black}{$+$1.12\,{\tiny($\pm$0.37)}} & \textcolor{green!60!black}{$+$0.73\,{\tiny($\pm$0.39)}} & $+$0.40\,{\tiny($\pm$0.51)} \\
\midrule
\multicolumn{5}{@{}l}{\emph{$k=4$}} \\
ZS & 74.65 & \textcolor{green!60!black}{$+$6.33\,{\tiny($\pm$0.91)}} & \textcolor{green!60!black}{$+$4.15\,{\tiny($\pm$0.77)}} & \textcolor{green!60!black}{$+$2.18\,{\tiny($\pm$0.50)}} \\
SimpleShot & 73.83 & \textcolor{green!60!black}{$+$8.48\,{\tiny($\pm$0.73)}} & \textcolor{green!60!black}{$+$7.74\,{\tiny($\pm$0.70)}} & \textcolor{green!60!black}{$+$0.75\,{\tiny($\pm$0.31)}} \\
Linear Probe & 75.84 & \textcolor{green!60!black}{$+$6.35\,{\tiny($\pm$0.66)}} & \textcolor{green!60!black}{$+$4.94\,{\tiny($\pm$0.68)}} & \textcolor{green!60!black}{$+$1.41\,{\tiny($\pm$0.34)}} \\
Tip-Adapter & 79.05 & \textcolor{green!60!black}{$+$2.16\,{\tiny($\pm$0.45)}} & \textcolor{green!60!black}{$+$1.44\,{\tiny($\pm$0.32)}} & \textcolor{green!60!black}{$+$0.72\,{\tiny($\pm$0.34)}} \\
TIP-X & 76.93 & \textcolor{green!60!black}{$+$0.52\,{\tiny($\pm$0.13)}} & \textcolor{green!60!black}{$+$0.46\,{\tiny($\pm$0.11)}} & $+$0.07\,{\tiny($\pm$0.11)} \\
APE & 79.78 & \textcolor{green!60!black}{$+$1.64\,{\tiny($\pm$0.36)}} & \textcolor{green!60!black}{$+$1.10\,{\tiny($\pm$0.30)}} & \textcolor{green!60!black}{$+$0.53\,{\tiny($\pm$0.30)}} \\
TAMP+ZS & 78.80 & \textcolor{green!60!black}{$+$2.20\,{\tiny($\pm$0.46)}} & -- & -- \\
LDA+ZS & 82.12 & \textcolor{green!60!black}{$+$0.68\,{\tiny($\pm$0.26)}} & $+$0.02\,{\tiny($\pm$0.26)} & \textcolor{green!60!black}{$+$0.65\,{\tiny($\pm$0.26)}} \\
\midrule
\multicolumn{5}{@{}l}{\emph{$k=8$}} \\
ZS & 74.65 & \textcolor{green!60!black}{$+$8.90\,{\tiny($\pm$1.10)}} & \textcolor{green!60!black}{$+$5.35\,{\tiny($\pm$0.88)}} & \textcolor{green!60!black}{$+$3.55\,{\tiny($\pm$0.58)}} \\
SimpleShot & 78.80 & \textcolor{green!60!black}{$+$5.54\,{\tiny($\pm$0.47)}} & \textcolor{green!60!black}{$+$4.66\,{\tiny($\pm$0.44)}} & \textcolor{green!60!black}{$+$0.88\,{\tiny($\pm$0.21)}} \\
Linear Probe & 81.44 & \textcolor{green!60!black}{$+$2.61\,{\tiny($\pm$0.36)}} & \textcolor{green!60!black}{$+$1.10\,{\tiny($\pm$0.39)}} & \textcolor{green!60!black}{$+$1.51\,{\tiny($\pm$0.28)}} \\
Tip-Adapter & 81.37 & \textcolor{green!60!black}{$+$2.79\,{\tiny($\pm$0.35)}} & \textcolor{green!60!black}{$+$1.02\,{\tiny($\pm$0.29)}} & \textcolor{green!60!black}{$+$1.77\,{\tiny($\pm$0.28)}} \\
TIP-X & 78.84 & $+$0.03\,{\tiny($\pm$0.06)} & \textcolor{green!60!black}{$+$0.19\,{\tiny($\pm$0.11)}} & \textcolor{red!80!black}{$-$0.16\,{\tiny($\pm$0.12)}} \\
APE & 82.06 & \textcolor{green!60!black}{$+$2.29\,{\tiny($\pm$0.38)}} & \textcolor{green!60!black}{$+$0.86\,{\tiny($\pm$0.24)}} & \textcolor{green!60!black}{$+$1.43\,{\tiny($\pm$0.27)}} \\
TAMP+ZS & 80.00 & \textcolor{green!60!black}{$+$3.30\,{\tiny($\pm$0.54)}} & -- & -- \\
LDA+ZS & 85.24 & \textcolor{green!60!black}{$+$0.49\,{\tiny($\pm$0.13)}} & $+$0.01\,{\tiny($\pm$0.09)} & \textcolor{green!60!black}{$+$0.48\,{\tiny($\pm$0.13)}} \\
\midrule
\multicolumn{5}{@{}l}{\emph{$k=16$}} \\
ZS & 74.65 & \textcolor{green!60!black}{$+$10.04\,{\tiny($\pm$1.20)}} & \textcolor{green!60!black}{$+$6.24\,{\tiny($\pm$1.01)}} & \textcolor{green!60!black}{$+$3.80\,{\tiny($\pm$0.59)}} \\
SimpleShot & 81.56 & \textcolor{green!60!black}{$+$4.00\,{\tiny($\pm$0.37)}} & \textcolor{green!60!black}{$+$3.01\,{\tiny($\pm$0.30)}} & \textcolor{green!60!black}{$+$0.99\,{\tiny($\pm$0.21)}} \\
Linear Probe & 85.12 & \textcolor{green!60!black}{$+$0.37\,{\tiny($\pm$0.27)}} & \textcolor{red!80!black}{$-$0.79\,{\tiny($\pm$0.32)}} & \textcolor{green!60!black}{$+$1.16\,{\tiny($\pm$0.20)}} \\
Tip-Adapter & 82.85 & \textcolor{green!60!black}{$+$2.69\,{\tiny($\pm$0.33)}} & \textcolor{green!60!black}{$+$1.02\,{\tiny($\pm$0.20)}} & \textcolor{green!60!black}{$+$1.67\,{\tiny($\pm$0.22)}} \\
TIP-X & 80.58 & \textcolor{green!60!black}{$+$0.43\,{\tiny($\pm$0.20)}} & \textcolor{green!60!black}{$+$0.25\,{\tiny($\pm$0.20)}} & $+$0.18\,{\tiny($\pm$0.21)} \\
APE & 83.84 & \textcolor{green!60!black}{$+$2.11\,{\tiny($\pm$0.35)}} & \textcolor{green!60!black}{$+$0.62\,{\tiny($\pm$0.17)}} & \textcolor{green!60!black}{$+$1.49\,{\tiny($\pm$0.23)}} \\
TAMP+ZS & 80.89 & \textcolor{green!60!black}{$+$3.78\,{\tiny($\pm$0.59)}} & -- & -- \\
LDA+ZS & 86.70 & \textcolor{green!60!black}{$+$0.70\,{\tiny($\pm$0.16)}} & \textcolor{green!60!black}{$+$0.36\,{\tiny($\pm$0.11)}} & \textcolor{green!60!black}{$+$0.34\,{\tiny($\pm$0.08)}} \\
\bottomrule
\end{tabular}
\end{table}

\newpage
\section{Hyperparameter Selection and Tuning}
\label{app:hyper}

This section reports our choices for choosing and searching hyperparameters of each method
to ensure complete transparency of our experiments.

\subsection{Hyperparameters for Each Method}
\label{app:hypermethod}

We first detail the hyperparameters used in each evaluated method and
their ensemble variants. For notational simplicity, we adopt a slightly different
convention here compared to Section~\ref{sec:clip}. Specifically, every scoring
function defined below operates on a single test image embedding $x \in \mathbb{R}^d$ and
yields a score vector $\mathrm{score}(x) \in \mathbb{R}^K$ across all $K$ classes. The $c$-th entry,
$\mathrm{score}(x)_c$, represents the logit for class $c$, such that the final predicted label
is given by $\hat{y} = \arg \max_c \mathrm{score}(x)_c$. 
Throughout the section, we will use $w$ to denote DEFTS's combining weight whenever DEFTS's score is added to an existing method.

First, we define the baseline zero-shot logit vector as $s_{\mathrm{ZS}}(x) \in \mathbb{R}^K$, with its per-class entry computed via cosine similarity:
\begin{align*}
    s_{\mathrm{ZS}}(x)_c = \frac{x}{\lVert x\rVert} \cdot \frac{t_c}{\lVert t_c\rVert},
\end{align*}
where $t_c$ represents the text embedding for class $c$.

\paragraph{Tip-Adapter and Variants.}
Tip-Adapter relies on a key-value cache built from the support set. The hyperparameter $\beta$ controls the sharpness of the affinity kernel, and $\alpha$ acts as an additive weight on the resulting cache vote. Let $\ell(x_j)$ denote the true label of support image $x_j$. For a given class $c$, the total affinity contributed by class-$c$ cache entries is $C(x)_c$:
\begin{align*}
A_j(x) &= \exp\big(-\beta(1 - \cos(x, x_j))\big), \qquad
C(x)_c = \sum_{j:\, \ell(x_j)=c} A_j(x) \\
\mathrm{score}(x)_c &= s_{\mathrm{ZS}}(x)_c + \alpha\, C(x)_c
\end{align*}

\textbf{TIP-X} introduces a third parameter $\gamma$, weighting a second, independent vote $D(x)_c$. This vote is built from a KL-divergence-based term $\psi_j(x)$ that measures how well cache entry $x_j$'s zero-shot distribution matches $x$'s:
\begin{align*}
D(x)_c &= \sum_{j:\, \ell(x_j)=c} \psi_j(x) \\
\mathrm{score}(x)_c &= s_{\mathrm{ZS}}(x)_c + \alpha\, C(x)_c + \gamma\, D(x)_c
\end{align*}

\textbf{APE} utilizes the same structural roles for $\beta, \alpha, \gamma$, but computes $A_j(x)$ within a channel-reduced feature space. Here $\gamma$ weights each cache entry's own reliability $r_j = \exp(\gamma \cdot \mathrm{CE}_j)$, where $\mathrm{CE}_j \geq 0$ is the cross-entropy of entry $j$'s reduced-space zero-shot prediction against its true label.
\begin{equation}
\mathrm{score}(x)_c = s_{\mathrm{ZS}}(x)_c + \alpha \sum_{j:\, \ell(x_j)=c} A_j(x)\, r_j
\end{equation}

\textbf{Augmenting with DEFTS.} DEFTS's own posterior score $s_{\mathrm{DEFTS}}(x)_c$ as computed in Algorithm~\ref{app:algorithm} is added as an additional term to each of the three formulas above, with weight $w$:
\begin{align*}
\text{\textbf{Tip-Adapter+DEFTS:}}\quad \mathrm{score}(x)_c &= s_{\mathrm{ZS}}(x)_c + \alpha\, C(x)_c + w\, s_{\mathrm{DEFTS}}(x)_c \\
\text{\textbf{TIP-X+DEFTS:}}\quad \mathrm{score}(x)_c &= s_{\mathrm{ZS}}(x)_c + \alpha\, C(x)_c + \gamma\, D(x)_c + w\, s_{\mathrm{DEFTS}}(x)_c \\
\text{\textbf{APE+DEFTS:}}\quad \mathrm{score}(x)_c &= s_{\mathrm{ZS}}(x)_c + \alpha \sum_{j:\, \ell(x_j)=c} A_j(x)\, r_j + w\, s_{\mathrm{DEFTS}}(x)_c
\end{align*}

\paragraph{LDA Ensembles.}
The Mahalanobis-based LDA score is defined per class as:
\begin{equation}
s_{\mathrm{LDA}}(x)_c = x^\top \hat{\Sigma}^{-1} \mu_c - \tfrac{1}{2} \mu_c^\top \hat{\Sigma}^{-1} \mu_c,
\end{equation}
with $\mu_c$ representing the class-$c$ support-image mean, and $\hat{\Sigma}^{-1}$ denoting the shrunk pooled precision matrix \citep{wang2024hardtobeat}. The hyperparameter $\beta$ controls the weights of the zero-shot score and this LDA score:
\begin{align*}
\text{\textbf{LDA+ZS:}} \quad \mathrm{score}(x)_c &= s_{\mathrm{ZS}}(x)_c + \beta\, s_{\mathrm{LDA}}(x)_c
\end{align*}
When combining with DEFTS, DEFTS's score replaces zero-shot's role in the same additive
structure, with $w$ taking over $\beta$'s share of the zero-shot-vs-DEFTS split:
\begin{align*}
\text{\textbf{LDA+ZS+DEFTS:}} \quad \mathrm{score}(x)_c &= (1-w)\, s_{\mathrm{ZS}}(x)_c + w\, s_{\mathrm{DEFTS}}(x)_c + \alpha\, s_{\mathrm{LDA}}(x)_c
\end{align*}

\paragraph{TAMP Ensembles.}
For the base TAMP method, $\lambda$ balances the image-derived mean $\mu_c$ and the text prototype $t_c$ (where $P$ is the text-aligned projector). Class $c$'s test-point score is the cosine similarity to this mixed prototype $W_c$:
\begin{align*}
W_c &= \lambda \Big(\frac{\mu_c}{\lVert\mu_c\rVert} P\Big) + (1-\lambda)\, t_c \\
s_{\mathrm{TAMP}}(x)_c &= \Big(\frac{x}{\lVert x\rVert} P\Big) \cdot W_c
\end{align*}
When ensembled with DEFTS, the scores are combined as follows:
\begin{align*}
\text{\textbf{TAMP+ZS:}} \quad \mathrm{score}(x)_c &= \beta\, s_{\mathrm{ZS}}(x)_c + (1-\beta)\, s_{\mathrm{TAMP}}(x)_c \\
\text{\textbf{TAMP+ZS+DEFTS:}} \quad \mathrm{score}(x)_c &= \beta\, s_{\mathrm{ZS}}(x)_c + (1-\beta)\big[w\, s_{\mathrm{DEFTS}}(x)_c + (1-w)\, s_{\mathrm{TAMP}}(x)_c\big]
\end{align*}

\paragraph{DEFTS Ensembles (Ours).}
DEFTS computes its own posterior-based score, $s_{\mathrm{DEFTS}}(x) \in \mathbb{R}^K$, without requiring internal hyperparameter tuning as in Algorithm~\ref{alg:defts}. Ensembled with zero-shot alone:
\begin{align*}
\text{\textbf{DEFTS+ZS:}} \quad \mathrm{score}(x)_c &= \alpha\, s_{\mathrm{ZS}}(x)_c + (1-\alpha)\, s_{\mathrm{DEFTS}}(x)_c.
\end{align*}

\paragraph{DEFTS Augmenting Simple Baselines.}
SimpleShot classifies by nearest class-mean cosine similarity, with no hyperparameters of its
own:
\begin{equation}
s_{\mathrm{SS}}(x)_c = \frac{x}{\lVert x\rVert} \cdot \frac{\mu_c}{\lVert \mu_c\rVert}.
\end{equation}
Linear Probe fits an $\ell_2$-regularized multinomial logistic regression on the $k$-shot
support embeddings and scores $x$ with its decision function,
$s_{\mathrm{LP}}(x)_c = \omega_c^\top (x/\lVert x\rVert) + b_c$, for fitted weights
$(\omega_c, b_c)$. Neither method otherwise touches the text embeddings at all -- DEFTS is the
only source of text information for both once combined:
\begin{align*}
\text{\textbf{SimpleShot+DEFTS:}} \quad \mathrm{score}(x)_c &= w\, s_{\mathrm{SS}}(x)_c + (1-w)\, s_{\mathrm{DEFTS}}(x)_c. \\
\text{\textbf{Linear Probe+DEFTS:}} \quad \mathrm{score}(x)_c &= w\, s_{\mathrm{LP}}(x)_c + (1-w)\, s_{\mathrm{DEFTS}}(x)_c.
\end{align*}

\subsection{Fixed Weights}
\label{sec:fixedweights}

When $k = 1$, i.e. each class has only one training sample, 
we choose fixed hyperparameters for all methods considered.

\paragraph{Natural untuned defaults for ensembling.} 
For ensembling different methods, we default to equal weighting (e.g., $\lambda=0.5$, $\beta=0.5$). This equal-weighting strategy serves as the most natural possible baseline without hyperparameter tuning.

\paragraph{Empirical defaults for LDA ensembles.} 
While equal weighting is the natural default for ensembling method, we observed a severe scale mismatch in the LDA formulation of \citet{wang2024hardtobeat}: their raw LDA scores are consistently $\sim\!10^3$ times larger in magnitude than the logits in zero-shot CLIP. Consequently, when ensembling LDA with other methods, we down scale the weight of LDA to $0.001$.

The exact fixed values used for every hyperparameter across all methods are summarized in Table~\ref{tab:fixed_hparams}.

\begin{table}[h]
\centering
\caption{Fixed hyperparameter values chosen for each method and its ensemble variants.}
\label{tab:fixed_hparams}
\begin{tabular}{l l l}
\toprule
\textbf{Method} & \textbf{Parameter(s)} & \textbf{Fixed value} \\
\midrule
Tip-Adapter & $(\beta,\ \alpha)$ & $(1.0,\ 1.0)$ \\
TIP-X & $(\beta,\ \alpha,\ \gamma)$ & $(1.0,\ 1.0,\ 1.0)$ \\
APE  & $(\beta,\ \alpha,\ \gamma)$ & $(1.0,\ 1.0,\ 0.1)$ \\
LDA + ZS & $\beta$ & $0.001$ \\
TAMP  & $\lambda$ & $0.5$ \\
TAMP + ZS & $(\lambda,\ \beta)$ & $(0.5,\ 0.5)$ \\
\midrule
ZS + DEFTS & $w$ & $0.5$ \\
SimpleShot + DEFTS & $w$ & $0.5$ \\
Linear Probe + DEFTS & $w$ & $0.5$ \\
Tip-Adapter + DEFTS & $(\beta,\ \alpha,\ w)$ & $(1.0,\ 1.0,\ 1.0)$ \\
TIP-X + DEFTS & $(\beta,\ \alpha,\ \gamma,\ w)$ & $(1.0,\ 1.0,\ 1.0,\ 1.0)$ \\
APE + DEFTS & $(\beta,\ \alpha,\ \gamma,\ w)$ & $(1.0,\ 1.0,\ 0.1,\ 1.0)$ \\
TAMP + ZS + DEFTS & $(\lambda,\ \beta,\ w)$ & $(0.5,\ 2/3,\ 0.5)$ \\
LDA + ZS + DEFTS & $(\alpha,\ w)$ & $(0.0005,\ 0.5)$ \\
\bottomrule
\end{tabular}
\end{table}

\subsection{Hyperparameter Search Methodology}
\label{sec:hyperparametersearch}

When $k \geq 2$, we perform a hyperparameter search for each method to identify its optimal configuration. For every method in our comparison, this search is performed via leave-one-out (LOO) cross-validation on the $k$-shot training set to find the hyperparameter combination that maximizes classification accuracy. Specifically, for each support image, the model is fit without that image and then used to score it. The hyperparameter combination yielding the highest LOO accuracy is selected. In the event of a tie, the configuration that minimizes the cross-entropy loss is chosen.

The search space for each hyperparameter of each method is shown in Table~\ref{tab:hparam_search}.

\begin{table}[h]
\centering
\caption{Hyperparameters searched via leave-one-out (LOO) cross-validation for each method and ensemble variant. All grids are defined below the table; ``combos'' is the total number of joint grid points evaluated for each method.}
\label{tab:hparam_search}
\begin{tabular}{l l l r}
\toprule
\textbf{Method} & \textbf{Parameter(s)} & \textbf{Grid(s)} & \textbf{combos} \\
\midrule
Tip-Adapter & $(\beta,\ \alpha)$ & $\mathcal{G}^{\mathrm{TIP}}_{\beta} \times \mathcal{G}^{\mathrm{TIP}}_{\alpha}$ & 72 \\
TIP-X & $(\beta,\ \alpha,\ \gamma)$ & $\mathcal{G}^{\mathrm{TIP}}_{\beta} \times \mathcal{G}^{\mathrm{TIP}}_{\alpha} \times \mathcal{G}^{\mathrm{TIPX}}_{\gamma}$ & 504 \\
APE & $(\beta,\ \alpha,\ \gamma)$ & $\mathcal{G}^{\mathrm{TIP}}_{\beta} \times \mathcal{G}^{\mathrm{TIP}}_{\alpha} \times \mathcal{G}^{\mathrm{APE}}_{\gamma}$ & 432 \\
LDA + ZS & $\alpha$ & $\mathcal{G}_{\alpha}$ & 9 \\
TAMP & $\lambda$ & $\mathcal{G}_{21}$ & 21 \\
TAMP+ZS & $(\lambda,\ \beta)$ & $\mathcal{G}_{21} \times \mathcal{G}_{21}$ & 441 \\
\midrule
ZS+DEFTS & $w$ & $\mathcal{G}_{21}$ & 21 \\
SimpleShot+DEFTS & $w$ & $\mathcal{G}_{21}$ & 21 \\
Linear Probe+DEFTS & $w$ & $\mathcal{G}_{21}$ & 21 \\
Tip-Adapter+DEFTS & $(\beta,\ \alpha,\ w)$ & $\mathcal{G}^{\mathrm{TIP}}_{\beta} \times \mathcal{G}^{\mathrm{TIP}}_{\alpha} \times \mathcal{G}^{\mathrm{DEFTS}}_{w}$ & 576 \\
TIP-X+DEFTS & $(\beta,\ \alpha,\ \gamma,\ w)$ & $\mathcal{G}^{\mathrm{TIP}}_{\beta} \times \mathcal{G}^{\mathrm{TIP}}_{\alpha} \times \mathcal{G}^{\mathrm{TIPX}}_{\gamma} \times \mathcal{G}^{\mathrm{DEFTS}}_{w}$ & 4032 \\
APE+DEFTS & $(\beta,\ \alpha,\ \gamma,\ w)$ & $\mathcal{G}^{\mathrm{TIP}}_{\beta} \times \mathcal{G}^{\mathrm{TIP}}_{\alpha} \times \mathcal{G}^{\mathrm{APE}}_{\gamma} \times \mathcal{G}^{\mathrm{DEFTS}}_{w}$ & 3456 \\
TAMP+ZS+DEFTS & $(\lambda,\ \beta,\ w)$ & $\mathcal{G}_{21} \times \mathcal{G}_{21} \times \mathcal{G}_{21}$ & 9261 \\
LDA+ZS+DEFTS & $(\alpha,\ w)$ & $\mathcal{G}_{\alpha} \times \mathcal{G}_{21}$ & 189 \\
\bottomrule
\end{tabular}

\vspace{0.75em}
\footnotesize
\begin{description}
    \item[$\mathcal{G}_{21}$:] $\{0, 0.05, 0.10, \dots, 1.0\}$ (21 points).
    \item[$\mathcal{G}_{\alpha}$:] $\{0, 0.05, 0.1, 0.35, 0.5, 1.0, 1.5, 2.5, 4.0\} \times 10^{-3}$ (9 points).
    \item[$\mathcal{G}^{\mathrm{TIP}}_{\beta}$:] $\{1,3,5,7,10,15,20,50\}$ (8 points).
    \item[$\mathcal{G}^{\mathrm{TIP}}_{\alpha}$:] $\{0.5,1,3,5,7,10,15,20,50\}$ (9 points).
    \item[$\mathcal{G}^{\mathrm{TIPX}}_{\gamma}$:] $\{0.5,2,7,20,50,100,200\}$ (7 points).
    \item[$\mathcal{G}^{\mathrm{APE}}_{\gamma}$:] $\{0, 0.1, 0.35, 0.75, 1.5, 3.0\}$ (6 points).
    \item[$\mathcal{G}^{\mathrm{DEFTS}}_{w}$:] $\{0, 0.5, 1.0, 1.5, 2.0, 2.5, 3.0, 3.5\}$ (8 points).
\end{description}
\vspace{0.25em}
\end{table}

\end{document}